%% file: iclr2026_conference.tex
\PassOptionsToPackage{table}{xcolor}
\documentclass{article} 
\let\OriginalAddContentsLine\addcontentsline
\usepackage{iclr2026_conference,times}
\let\addcontentsline\OriginalAddContentsLine

\input{math_commands.tex}

\usepackage{url}
\usepackage{booktabs}
\usepackage{tabularx}
\usepackage{graphicx}
\usepackage{multirow}
\usepackage{xcolor}
\usepackage{amsmath}
\usepackage{amssymb}
\usepackage{algorithm}
\usepackage{algpseudocode}

\usepackage{amsmath,amssymb}
\usepackage{algorithm}
\usepackage{algpseudocode}
\usepackage[normalem]{ulem}
\usepackage{placeins}

\usepackage{xspace}
\usepackage{booktabs}
\usepackage{graphicx}
\usepackage{pifont}
\usepackage[table]{xcolor}

\usepackage{booktabs}
\usepackage{caption}
\usepackage{setspace}

\usepackage{array}
\newlength{\authorcolwidth}

\newcommand{\cmark}{\ding{51}}
\newcommand{\xmark}{\textcolor{gray!65}{\ding{55}}}

\definecolor{oursbg}{RGB}{239,245,252}

\newcommand{\maintablefont}{\fontsize{8}{9}\selectfont}

\usepackage{xcolor}

\newcommand{\ourmethod}{\textsc{AutoAgent0}\xspace}

\usepackage{pifont}

\definecolor{myiclrblue}{RGB}{152, 182, 231}
\definecolor{darkgreen}{RGB}{0, 154, 85}
\newcommand{\avggain}[1]{\,{\fontsize{5.5}{6.5}\selectfont\textcolor{darkgreen}{+#1}}}
\usepackage[colorlinks=true,
            linkcolor=myiclrblue,      
            citecolor=myiclrblue,  
            urlcolor=myiclrblue]{hyperref}
\usepackage{cleveref}

\title{\raisebox{-.18\height}{\includegraphics[width=0.07\textwidth]{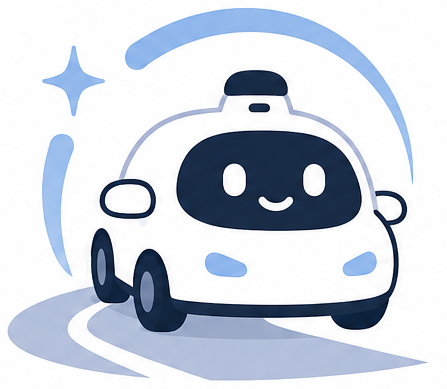}} \hspace{0.1em}%
  \textcolor{myiclrblue}{\textbf{AutoAgent0}}: An Agentic Runtime for Safe Closed-loop Driving \vspace{-0.33cm}}

\author{%
\setlength{\authorcolwidth}{0.26\textwidth}%
\setlength{\tabcolsep}{0pt}%
\begin{tabular}[t]{@{}*{4}{>{\raggedright\arraybackslash}p{\authorcolwidth}}@{}}
\textbf{Jiageng Liu}$^1$ &
\textbf{Seth Z. Zhao}$^1$\thanks{Project leads and corresponding authors:
\texttt{sethzhao506@g.ucla.edu},
\texttt{zeweizhou@ucla.edu}.} &
\textbf{Zewei Zhou}$^1$\footnotemark[1] &
\textbf{Alejandro Huerta}$^1$ \\[1pt]
\textbf{Aidan Bayer-Calvert}$^1$ &
\textbf{Yuxin Bao}$^1$ &
\textbf{Haotian Lin}$^2$ &
\textbf{Robert Spataru}$^1$ \\[1pt]
\textbf{Zijun Cui}$^3$ &
\textbf{Yining Hong}$^4$ &
\textbf{Bolei Zhou}$^1$ &
\textbf{Jiaqi Ma}$^1$
\end{tabular}\\
\noalign{\vskip 4pt}
{\normalfont
$^1$UCLA\hspace{1.5em}%
$^2$CMU\hspace{1.5em}%
$^3$UT Dallas\hspace{1.5em}%
$^4$Stanford
}\hspace{1.5em}%
{\normalfont\scriptsize\ttfamily
\href{https://jiagengliu02.github.io/AutoAgent0/}%
{https://jiagengliu02.github.io/AutoAgent0/}%
}
}

\usepackage{pgfplots}
\usepgfplotslibrary{groupplots}
\pgfplotsset{compat=1.18}

\iclrfinalcopy 
\begin{document}

\maketitle

\input{sec/abstract}
\input{sec/introduction}

\input{sec/related_work}

\input{sec/method}

\input{sec/experiments}
\input{sec/discussion_and_limitations}


\clearpage



\bibliography{iclr2026_conference}
\bibliographystyle{iclr2026_conference}

\appendix
\input{sec/appendix}

\end{document}

%% file: math_commands.tex
\usepackage{amsmath,amsfonts,bm}

\def\eqref#1{equation~\ref{#1}}

\def\1{\bm{1}}

\DeclareMathAlphabet{\mathsfit}{\encodingdefault}{\sfdefault}{m}{sl}
\SetMathAlphabet{\mathsfit}{bold}{\encodingdefault}{\sfdefault}{bx}{n}



%% file: sec/abstract.tex
\begin{figure}[h]
  \centering
  \vspace{-0.5cm}%
  \includegraphics[width=0.95\linewidth]{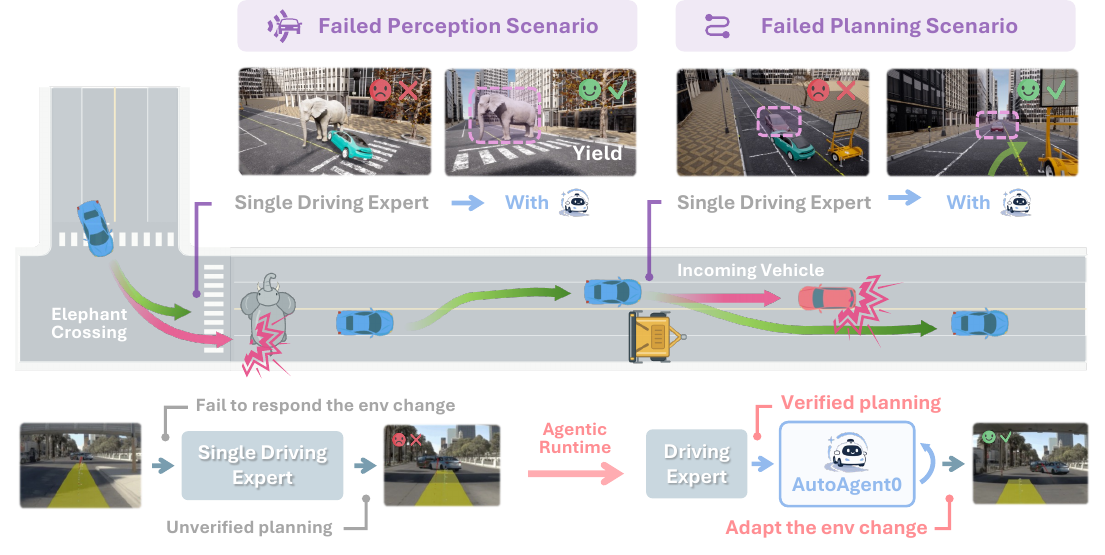}
  \vspace{-0.2cm}%
\caption{\textbf{Motivation and overview.}
Inaccurate scene context and limited generalization to off-nominal states motivate runtime supervision.
\ourmethod combines scene refinement, trajectory verification, and recovery planning.
Top: a Fail2Drive case illustrates yielding and bypass recovery.
Bottom: a NavSafe-$\infty$ example compares standalone and \ourmethod-augmented expert.}
  \label{fig1:teaser}
  \vspace{-0.1cm}%
\end{figure}

\begin{abstract}
Safe autonomous driving requires more than accurate planning at a single frame: it demands reliable closed-loop execution in dynamic environments.
Although end-to-end driving policies have advanced rapidly, the accumulated closed-loop errors, especially in unseen scenarios, can push vehicles into states from which the same policy struggles to recover.
We address the challenge by treating closed-loop driving safety as a test-time execution problem.
We introduce \textbf{\ourmethod}, an agentic runtime and harness framework that augments existing driving experts with complementary detection skills, rule-based action verification, and feedback-driven recovery.
To address inaccurate scene perception, an independent detection module constructs scene context using complementary perception skills, and a rule-based verifier checks proposed trajectories against this context before execution.
When the nominal trajectories are unreliable, the orchestrator of the runtime activates \textbf{Recovery Planning}, where an agentic loop constructs alternative action proposals from action primitives and iteratively refines them using verification feedback.
This design combines independent scene assessment with feasibility checks and online recovery, addressing both perception errors and the limited recovery capabilities of individual experts.
Evaluations on Fail2Drive and NavSafe-$\infty$ demonstrate improved closed-loop robustness over single-expert execution in long-tail and safety-critical scenarios.
These results demonstrate that runtime supervision can improve closed-loop robustness across multiple driving experts without retraining the underlying policies.
\end{abstract}

%% file: sec/introduction.tex
\section{Introduction}
\label{sec:intro}

Autonomous driving, a core application of physical AI \citep{agarwal2026cosmos,ali2025world}, operates in dynamic, open-ended environments and imposes stringent requirements for closed-loop safety and generalizability. The field has progressed from modular pipelines \citep{li2024bevformer,liu2025hybrid} toward end-to-end driving systems \citep{uniad,zhou2025v2xpnp,zhou2025turbotrain}. Nevertheless, safe closed-loop driving remains challenging: real-world scenarios follow a long-tailed distribution that cannot be exhaustively covered during training, while small decision errors can compound through interaction and eventually lead to safety-critical failures.

The dominant paradigm trains a single expert policy to directly produce actions across diverse scenarios \citep{egostatus,xu2026wod}. Despite this progress, such policies often generalize poorly to unfamiliar long-tail events in closed-loop driving \citep{bench2drive,zhao2026bridgesim}. The accumulated errors may leave the vehicle stuck and unable to be recovered by the same policy that caused the deviation. For example, Fail2Drive \citep{fail2drive} reports an average performance drop of 16.3\% across seven driving models under long-tail closed-loop evaluation. Moreover, single learning-based experts remain largely opaque, making their decisions difficult to trust across diverse long-tail scenarios \citep{Kim_2026_CVPR}. These limitations raise a central question: \textit{\uline{Should safe long-tail driving rely solely on an increasingly capable expert policy, or should safety also be addressed through explicit runtime supervision at test time?}}

Beyond generating high-quality actions, safe closed-loop driving requires a system to detect failures, propose alternatives, verify their feasibility, and recover online \citep{karkus2025beyond}. 
Recent agentic systems have demonstrated strong closed-loop execution capabilities by coordinating specialized skills, inspecting execution traces, analyzing failures, and recovering through feedback \citep{anthropic2025claudecode,liu2025robopilot,zhang2026qwen,zhou2026verifine}. 
This progress motivates an agentic approach to driving that combines the reasoning capabilities of vision language models (VLMs) with the basic driving capabilities of existing experts. Rather than relying on a single expert in all situations, a runtime can orchestrate these complementary capabilities, verify proposed actions, and initiate recovery when the expert fails to maintain safe progress under changing scene conditions.

To this end, we introduce \textbf{\ourmethod}, an agentic runtime for safe closed-loop driving, as illustrated in \Cref{fig:autoagent0-runtime}. \ourmethod is motivated by two sources of failure in end-to-end driving: inaccurate scene perception and limited capability to recover when single-expert execution breaks down, as shown in \Cref{fig1:teaser}. \ourmethod addresses these limitations through a runtime that integrates complementary detection skills, rule-based action verification, and feedback-driven recovery. First, for robust and interpretable verification before execution, an independent detection module constructs scene context using complementary detection tools. Second, a rule-based verifier checks action feasibility given the detected scene. When the nominal trajectory from the existing driving policy is deemed unreliable, the orchestrator activates \textbf{Recovery Planning}, where an agentic loop then constructs alternative maneuvers from action primitives and iteratively refines them using feedback from the rule-based verifier to restore feasible driving behavior.

We evaluate \ourmethod on challenging closed-loop benchmarks, including Fail2Drive~\citep{fail2drive} and NavSafe-$\infty$~\citep{bao2026Navsafe}, with an emphasis on long-tail scenarios. The results demonstrate that runtime verification and feedback-driven recovery substantially improve robustness over single-expert execution. Our contributions are summarized as follows:
\begin{itemize}
\item We introduce \textbf{\ourmethod}, an agentic runtime that augments existing policies with complementary detection skills, rule-based action verification, and feedback-driven recovery.
\item We introduce an independent rule-based verifier for each trajectory before execution, while the \textbf{Recovery Planning} constructs and iteratively refines recovery maneuvers from action primitives when the nominal trajectory from the existing policy is unreliable.
\item We evaluate \ourmethod on Fail2Drive and NavSafe-$\infty$, demonstrating improved closed-loop robustness over single-expert execution in long-tail scenarios.
\end{itemize}

%% file: sec/related_work.tex
\section{Related Work}
\label{sec:related_work}

\paragraph{Closed-Loop Autonomous Driving}
Long-tail scenarios remain challenging for autonomous driving because rare interactions are underrepresented during training but can dominate closed-loop safety. Recent approaches include structured planners such as UniAD and VAD~\citep{uniad,vad}, multimodal and diffusion planners such as Hydra-MDP, DiffusionDrive, and SparseDriveV2~\citep{hydramdp,diffusiondrive,sun2026sparsedrivev2,zou2025diffusiondrivev2}, and VLA policies or world model policies such as AutoVLA and DriveVLA-W0~\citep{zhou2026autovla,zhou2026spanvla,li2025drivevla}. BridgeSim studies the open-loop to closed-loop gap, while Fail2Drive demonstrates persistent degradation on matched long-tail routes~\citep{zhao2026bridgesim,fail2drive}. 
Bench2Drive provides additional interactive evaluations~\citep{bench2drive}. World Engine improves the expert through post-training on reconstructed critical interactions~\citep{worldengine}. 
\ourmethod instead addresses long-tail failures at runtime without retraining the expert.

\paragraph{Closed-Loop Runtime Monitoring} Runtime supervision combines evidence for intervention with a corrective response. 
RSS defines safe-distance rules and proper responses~\citep{shalevshwartz2017rss}, while predictive safety filters optimize corrective controls under modeled dynamics and safety constraints~\citep{mpcshield}. 
Fallback-Safe MPC couples a calibrated perception monitor with planning that maintains fallback feasibility~\citep{sinha2023fallback}; 
SODA-MPC uses conformally calibrated ensemble uncertainty to switch to reachability-based control~\citep{contreras2025soda}. 
Recent studies further investigate language-model-assisted intervention selection~\citep{sinha2024realtimeanomalydetectionreactive} and plan repair under perception failures~\citep{chakraborty2024systemlevelsafetymonitoringrecovery}.
Building on these approaches, \ourmethod integrates scene-evidence refinement with parameterized recovery around a frozen driving expert.
Scene refinement updates the verifier's inputs, while recovery proposals are revised using verifier and execution feedback.

\paragraph{Agentic Runtime Harnesses}
Agentic runtimes coordinate tools, retain interaction state, and revise decisions using environmental feedback. 
ReAct and Toolformer establish reasoning-and-action and tool-use interfaces~\citep{yao2023react,schick2023toolformer}, while AutoGen and SWE-agent provide programmable execution harnesses ~\citep{wu2024autogen,yang2024sweagent}. 
Embodied systems such as Voyager, ENPIRE, and ASPIRE use execution feedback to explore and refine reusable skills~\citep{wang2024voyager,enpire,aspire}, and Agent-Driver applies tools, memory, and reasoning to autonomous driving~\citep{mao2024agentdriver}. 
Our harness applies these tool-use and memory mechanisms to driving recovery, with a rule-based verifier determining which proposed actions may reach the controller.

%% file: sec/method.tex
\section{\ourmethod: A Runtime Harness for Closed-Loop Driving}
\label{sec:method}

\subsection{Problem Setting and Closed-loop Runtime Formulation}
\label{sec:execution-loop}
\paragraph{Problem Setting.} We formulate end-to-end closed-loop driving as a route-conditioned POMDP process following prior works~\citep{fail2drive, bench2drive}. The original trajectory planner defines the default Nominal Policy $\pi_{\mathrm N}$, using learning-based or rule-based trajectory planners. At decision point $t$, this policy generates finite-horizon trajectory proposals $\tau\sim\pi_{\mathrm N}(\cdot\mid h_t,q)$.
Here $h_t=(o_0,\tau_0,\ldots,o_t)$ denotes the history of observations and previously executed trajectories, with $o_t$ the current observation.
The objective is to navigate to the goal region specified by the route instruction $q$ while maintaining safe execution throughout the closed-loop interaction.

\paragraph{Limitations of Nominal Planning.}
End-to-end nominal policies face two challenges in closed-loop driving.
\textit{(1) Inaccurate scene context.}
Their internal representations may miss safety-critical actors or mischaracterize drivable space, leading to unsafe trajectory proposals.
\textit{(2) Limited generalization to off-nominal states.}
Unexpected events or accumulated execution errors may lead to states poorly represented in training, where the policy may repeatedly propose unsafe trajectories or fail to make progress.
These challenges motivate a runtime that refines scene context to support trajectory verification and invokes recovery planning when rejecting nominal proposals alone cannot restore safe progress.


\begin{figure*}[t]
    \centering
    \includegraphics[width=\textwidth]{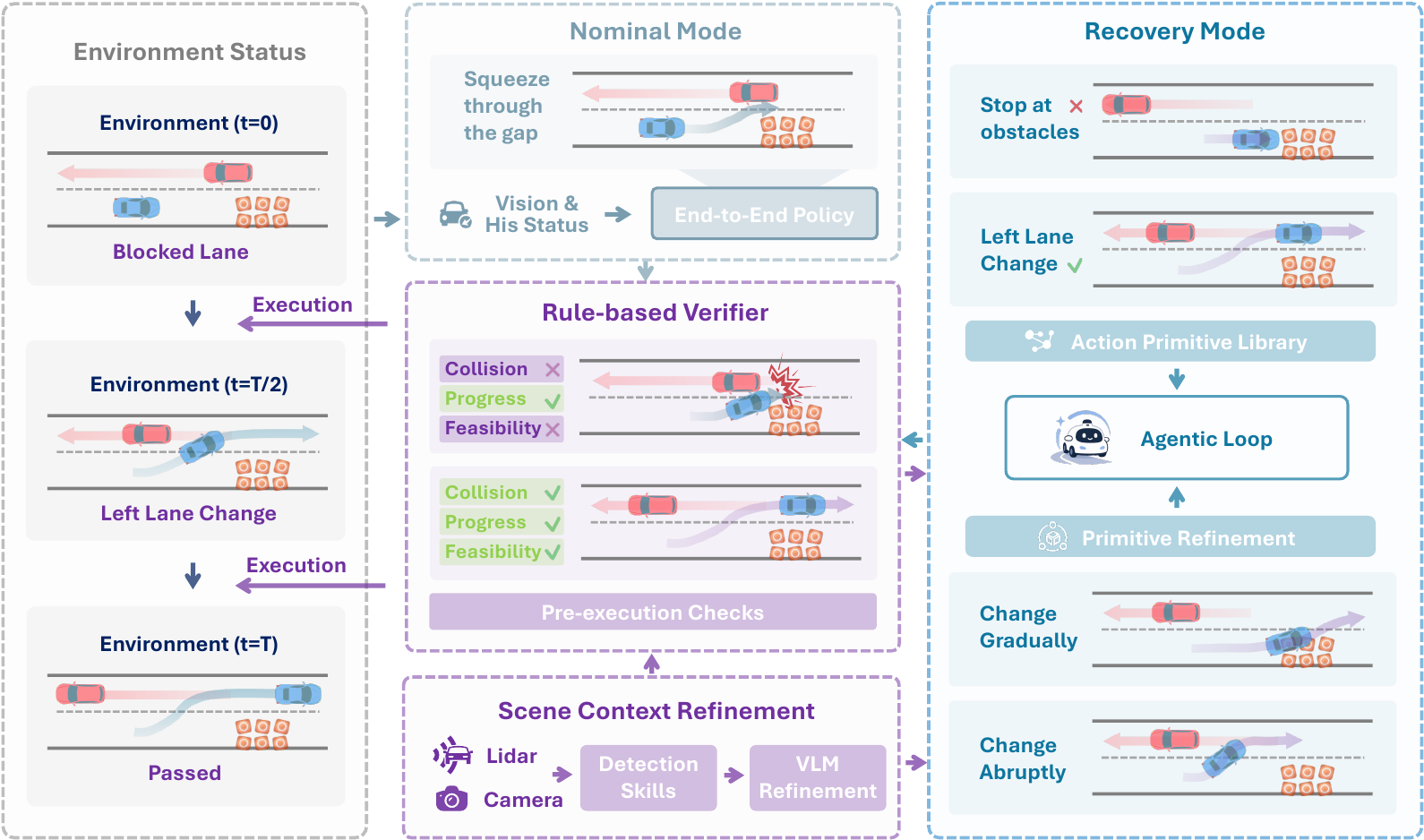}
    \caption{\textbf{Overview of the \ourmethod runtime.}
    \ourmethod surrounds the nominal end-to-end policy with a) scene context refinement, which uses perception skills to augment the evidence available to a rule-based verifier, and b) agentic recovery planning, which generates and revises recovery planning trajectories when no nominal proposal is admitted.
    The verifier evaluates both nominal proposals and recovery planning trajectories before commitment to the tracking controller.}
    \label{fig:autoagent0-runtime}
\end{figure*}

\paragraph{Closed-loop Runtime Formulation.}
A closed-loop runtime surrounds the nominal policy with a safety-enforcement layer comprising two complementary mechanisms~\citep{sinha2023fallback,mpcshield}.
\textit{(1) Proposal verification in nominal planning.}
Given runtime context $c_t$, comprising the observation history, route instruction, and runtime feedback, a verifier $V$ checks nominal proposals against specified safety criteria.
We denote the verified nominal policy by $\pi_{\mathrm N}^{V}$, which selects the highest-ranked admitted proposal according to the planning objective.
\textit{(2) Recovery planning.}
When no nominal proposal is admitted, the runtime invokes a recovery policy $\pi_{\mathrm R}$. Thus the executed trajectory $\tau_t$ at decision step $t$ could be written as:
\begin{equation}
    \tau_t=\operatorname{Runtime}(c_t)
    =
    \begin{cases}
        \pi_{\mathrm N}^{V}(c_t),
        & \text{if a nominal proposal is admitted},\\
        \pi_{\mathrm R}(c_t),
        & \text{otherwise}.
    \end{cases}
    \label{eq:runtime}
\end{equation}

\subsection{\ourmethod Runtime}
\label{sec:orchestrator-routing}

\ourmethod comprises a runtime harness that coordinates scene context refinement, trajectory verification, and recovery planning.
The harness invokes VLM reasoning for semantic assessment and recovery maneuver selection, integrating these judgments with perception outputs, verification results, and execution feedback.
Figure~\ref{fig:autoagent0-runtime} illustrates this structure; Section~\ref{sec:test-time-recovery} details the recovery planner.

\paragraph{Agent as the Runtime Operator.} Following the agentic operator formulation of AVO~\citep{chen2026avo}, the agent implements Eq.~\eqref{eq:runtime} using scene-refinement skills $\mathcal F$, parameterized recovery primitives $\mathcal P$, and a rule-based verifier $V$:
\begin{equation}
    \operatorname{Runtime}_{\mathrm{agent}}(c_t)
    =\operatorname{Agent}(c_t;\mathcal F,\mathcal P,V).
    \label{eq:runtime-agent}
\end{equation}
Here, the agent denotes the runtime harness, which invokes VLM reasoning for semantic assessment and recovery decisions.

\paragraph{Scene Context Refinement.} Reliable verification requires scene evidence that can expose errors in nominal proposals. 
The agent applies perception skills from $\mathcal F$ to remove false positive and false negative detections in $c_t$, producing refined context $\widetilde c_t$ for verification and recovery planning. 
The skills include cross-modal checks that identify discrepancies between camera and LiDAR observations, and vision-language reasoning provides semantic validation of relevant objects and scene conditions. 
These operations revise the scene representation used for verification, addressing missed actors and spurious detections that can otherwise cause incorrect admission of nominal proposals or unnecessary activation of recovery planning.

\paragraph{Rule-based Verification.}
The rule-based verifier $V$ evaluates nominal proposals and recovery planning trajectories against $\widetilde c_t$, returning an admission decision and feedback on violated criteria.
Its fixed admission criteria combine a threshold on a PDMS-based trajectory score~\citep{pdmclosed} with route-consistency and execution guards.
The score assesses collision and drivable-area compliance, time to collision, lane keeping, and ego progress; the guards account for conditions such as persistent blockage and confirmed semantic inconsistency.
When no nominal proposal is admitted, the agent implements $\pi_{\mathrm R}$ by generating recovery planning trajectories from $\mathcal P$ and using verifier feedback to revise primitive parameters or select another primitive.
Each revision is checked against $\widetilde c_t$ under the same admission criteria. The complete scoring function, hard gates, and admission threshold
are specified in Appendix~\ref{app:rules}. 
Section~\ref{sec:test-time-recovery} specifies this iterative refinement loop.

\subsection{Recovery Planning via Agentic Parameter Search}
\label{sec:test-time-recovery}

Recovery from off-nominal states requires corrective actions that the nominal policy, trained on largely nominal demonstrations, may fail to propose. 
\ourmethod reduces this dependence by decomposing recovery into components that do not rely on the expert's training data:
parameterized primitives generate trajectories analytically from the current geometry, 
the rule-based verifier checks them against explicit criteria, 
and the VLM contributes only high-level maneuver selection, which draws on broad pretrained reasoning rather than driving demonstrations. 
Moreover, recovery does not need to be correct in a single pass: rejection feedback allows the runtime to revise parameters or reconsider the maneuver before execution. The resulting generalization is bounded by the coverage of the primitive library (Section~\ref{app_sec:recovery-primitives}).

\paragraph{Action Primitive Library.} We define an action primitive library $\mathcal P$, where each primitive $\ell\in\mathcal P$ is associated with a trajectory generator $G_\ell$. A parameter vector $\alpha\in\mathcal A_\ell$ controls quantities such as lateral displacement, target speed, and execution duration, with $\mathcal A_\ell$ specifying the primitive's bounded parameter domain. For example, braking and creeping adjust longitudinal motion, lateral bypass changes the path around an obstacle, and reverse actions create space for subsequent forward progress. These primitives initially generate recovery planning trajectories that adapt to local geometry and route requirements through continuous parameter selection.

\begin{figure*}[t]
    \centering
    \includegraphics[width=0.8\textwidth]{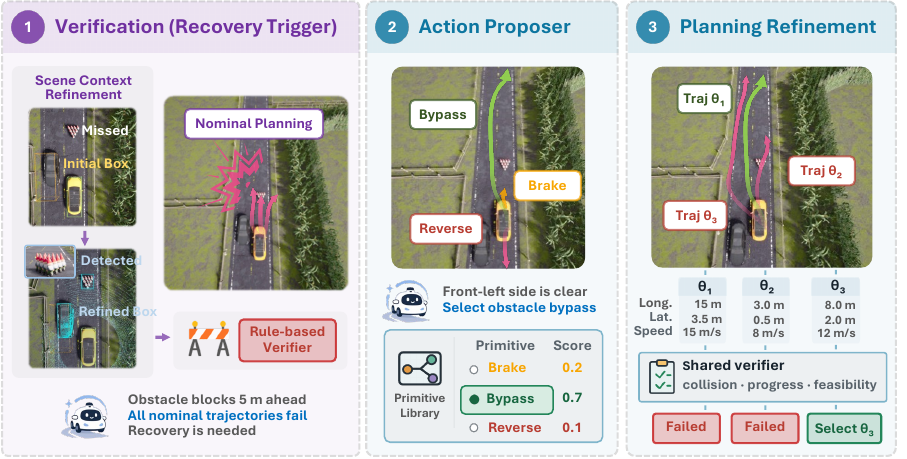}
    \caption{\textbf{Agentic Recovery Planning.}
    Rejection of nominal proposals initiates action primitive selection and parameterized trajectory generation.
    The agent uses scene context and verifier feedback to revise recovery planning trajectories before execution.
    During execution, new observations can trigger further revision, with each replacement trajectory checked before commitment.}
    \label{fig:recovery_mode}
\end{figure*}

\paragraph{Verifier-Guided Planning Refinement.}
The recovery policy $\pi_{\mathrm R}$ alternates primitive selection, trajectory generation, and verification (Figure~\ref{fig:recovery_mode}).
At each iteration, the agent selects a primitive $\ell$ and parameters $\alpha$ to generate a recovery planning trajectory:
    $\tau_{\text{recovery}}=G_\ell(\widetilde c_t,\alpha).
    \label{eq:recovery-proposal}$
Starting from nominal rejection feedback, the agent reasons about scene constraints and why proposals failed to decide whether to revise the current primitive’s parameters or select a different maneuver. 
It explores a parameter set for the selected primitive and evaluates candidates with the verifier $V$, retaining proposals and feedback for subsequent decisions. 
For example, the agent may infer from collision feedback and surrounding geometry that forward is constrained, and choose to reverse to create room for a subsequent turn.
Algorithm~\ref{alg:harness} provides the complete  procedure. 
Since recovery planning is triggered only when no nominal proposal is admitted, VLM reasoning is reserved for situations the nominal policy cannot resolve, rather than being invoked at every control step.
Appendix~\ref{app:runtime} reports the full latency analysis.

\vspace{-2mm}

%% file: sec/experiments.tex
\section{Experiments}
\label{sec:exp}

\vspace{-2mm}
\subsection{Experimental Protocol}
\label{sec:exp-protocol}

We evaluate whether \ourmethod serves as a reliable runtime for safe closed-loop driving without retraining the underlying driving expert. Our experiments mainly aim to answer the following research questions: 
\textbf{RQ1:} Can our runtime improve the closed-loop robustness of existing driving experts in long-tailed and safety-critical scenarios without retraining them?
\textbf{RQ2:} 
How do key harness components and VLM backbone choice affect the effectiveness of verification and recovery planning in \ourmethod?
and \textbf{RQ3:} How does our agentic runtime compare with alternative runtime methods in maintaining safety
and restoring progress?

\paragraph{Benchmarks and evaluation settings.} We mainly conduct experiments in Fail2Drive \citep{fail2drive}, which is a CARLA-based closed-loop benchmark that tests driving abilities under long-tailed scenarios. 
Additionally, we show the generalizability of \ourmethod in NavSafe-$\infty$~\citep{bao2026Navsafe}, which is photorealistic closed-loop benchmark that evaluates driving ability across safety-critical events. 
The evaluation metrics include success rate (SR), driving score (DS), along with HM on Fail2Drive and driving efficiency, comfort, and capability-specific scores on NavSafe-$\infty$. Detailed metric definitions are provided in Appendix.  
Closed-loop results are averaged over three matched evaluation repetitions unless otherwise stated. The runtime-monitor evaluation in Section~\ref{sec:rq3} uses 160 non-calibration routes, comprising 80 In-Distribution/Generalization pairs.

\paragraph{Implementation details.} Unless stated otherwise, we use SparseDriveV2~\citep{sun2026sparsedrivev2} as the driving expert, BridgeDrive~\citep{liu2025bridgedrive} as the sensor-based detector, and Opus 5~\citep{anthropic2026opus5}  for semantic supervision and recovery selection. Recovery variants use the same front-camera input to the VLM and the same deterministic primitive runtime. 
For Fail2Drive, we augment the frozen BridgeDrive detector with trained pedestrian and static-obstacle modules while keeping the driving expert frozen; configuration and training details are provided in Appendix~\ref{app:models-runtime-config}. NavSafe-$\infty$ uses native BridgeDrive perception.

\vspace{-2mm}
\subsection{Main results (RQ1)}
\label{sec:main-results}

\textbf{Results on Fail2Drive.}
Table~\ref{tab:main-results} compares \ourmethod-augmented UniAD, SparseDriveV2, and BridgeDrive with their standalone counterparts, alongside representative driving baselines.
\ourmethod consistently improves DS, SR, and HM across all three experts on both splits, with larger DS gains on generalization routes.
The gains mainly come from avoided collisions: most newly successful routes previously ended in collisions, and pedestrian and animal collisions are nearly eliminated.
Because standalone experts collide more often on generalization routes, removing these failures yields larger gains under distribution shift.
Some avoided collisions instead end in stops when no admissible recovery maneuver is found, reflecting the runtime's fail-safe preference and the coverage limits of the primitive library (Appendix~\ref{app_sec:scenario_breakdown_and_generalization}).

\begin{table*}[t]
\centering
\caption{
\textbf{Benchmark Results on Fail2Drive.}
HM is the harmonic mean of DS and SR within each split; Average is the arithmetic mean of the two splits for each metric.
Parentheses in Average indicate absolute improvements over the corresponding
base method.
}
\label{tab:main-results}
\vspace{0.1cm}
\maintablefont
\renewcommand{\arraystretch}{1.05}
\setlength{\tabcolsep}{1.75pt}

\begin{tabular*}{\textwidth}{@{\extracolsep{\fill}}>{\raggedright\arraybackslash}p{110pt} cc ccc ccc | r@{}l r@{}l r@{}l@{}}
\toprule
& \multicolumn{2}{c}{\textbf{Inputs}}
& \multicolumn{3}{c}{\textbf{In-Distribution}}
& \multicolumn{3}{c}{\textbf{Generalization}}
& \multicolumn{6}{c}{\textbf{Average}} \\
\cmidrule(lr){2-3}
\cmidrule(lr){4-6}
\cmidrule(lr){7-9}
\cmidrule(l){10-15}
\textbf{Method}
& RGB & LiDAR
& DS $\uparrow$ & \shortstack{SR\\(\%) $\uparrow$} & HM $\uparrow$
& DS $\uparrow$ & \shortstack{SR\\(\%) $\uparrow$} & HM $\uparrow$
& \multicolumn{2}{c}{DS $\uparrow$}
& \multicolumn{2}{c}{SR (\%) $\uparrow$}
& \multicolumn{2}{c}{HM $\uparrow$} \\
\midrule

TCP \citep{wu2022trajectory}
& \cmark & \xmark
& 24.7 & 39.1 & 30.3
& 24.5 & 31.4 & 27.5
& 24.6 & & 35.3 & & 28.9 & \\

Orion \citep{fu2025orion}
& \cmark & \xmark
& 53.0 & 52.0 & 52.5
& 51.2 & 46.0 & 48.5
& 52.1 & & 49.0 & & 50.5 & \\

HiP-AD \citep{tang2025hipad}
& \cmark & \xmark
& 74.1 & 70.7 & 72.4
& 67.1 & 56.7 & 61.5
& 70.6 & & 63.7 & & 67.0 & \\

SimLingo \citep{renz2025simlingo}
& \cmark & \xmark
& 82.6 & 79.3 & 80.9
& 71.7 & 55.0 & 62.2
& 77.2 & & 67.2 & & 71.6 & \\

AlignDrive \citep{wu2026aligndrive}
& \cmark & \xmark
& 74.7 & 75.7 & 75.2
& 68.6 & 62.7 & 65.5
& 71.7 & & 69.2 & & 70.4 & \\

BevAD \citep{holtz2026bevad}
& \cmark & \xmark
& 87.4 & 83.3 & 85.3
& 82.3 & 68.7 & 74.9
& 84.9 & & 76.0 & & 80.1 & \\

TF++ \citep{jaeger2023hidden}
& \cmark & \cmark
& 83.3 & 78.5 & 80.8
& 75.4 & 61.1 & 67.5
& 79.4 & & 69.8 & & 74.2 & \\

TFv6 \citep{lead2025}
& \cmark & \cmark
& 90.2 & 93.3 & 91.7
& 79.5 & 70.7 & 74.8
& 84.9 & & 82.0 & & 83.3 & \\

\midrule

UniAD \citep{uniad}
& \cmark & \xmark
& 47.5 & 36.3 & 41.2
& 44.0 & 27.6 & 33.9
& 45.8 & & 32.0 & & 37.6 & \\

\textbf{+ AutoAgent0}
\raisebox{-.18\height}{\includegraphics[width=1.4em]{fig/autoagent0_logo.png}}
& \cmark & \cmark
& 68.4 & 69.0 & 68.7
& 69.2 & 69.0 & 69.1
& 68.8 & \avggain{23.0}
& 69.0 & \avggain{37.0}
& 68.9 & \avggain{31.3} \\

\midrule

SparseDriveV2 \citep{sun2026sparsedrivev2}
& \cmark & \xmark
& 70.0 & 74.0 & 71.9
& 62.9 & 52.7 & 57.4
& 66.5 & & 63.4 & & 64.7 & \\

\textbf{+ AutoAgent0}
\raisebox{-.18\height}{\includegraphics[width=1.4em]{fig/autoagent0_logo.png}}
& \cmark & \cmark
& 80.7 & 85.7 & 83.1
& 79.2 & 81.0 & 80.1
& 80.0 & \avggain{13.5}
& 83.4 & \avggain{20.0}
& 81.7 & \avggain{17.0} \\

\midrule

BridgeDrive \citep{liu2025bridgedrive}
& \cmark & \cmark
& 91.6 & 95.0 & 93.3
& 81.9 & 75.0 & 78.3
& 86.8 & & 85.0 & & 85.8 & \\

\textbf{+ AutoAgent0}
\raisebox{-.18\height}{\includegraphics[width=1.4em]{fig/autoagent0_logo.png}}
& \textbf{\cmark} & \textbf{\cmark}
& \textbf{91.7} & \textbf{95.3} & \textbf{93.5}
& \textbf{85.7} & \textbf{85.3} & \textbf{85.5}
& \textbf{88.7} & \avggain{1.9}
& \textbf{90.3} & \avggain{5.3}
& \textbf{89.5} & \avggain{3.7} \\

\bottomrule
\end{tabular*}
\end{table*}

\textbf{Results on NavSafe-$\infty$.}
To show the generalizability of \ourmethod, we compare the performance enhancement in photorealistic closed-loop benchmark NavSafe-$\infty$ (Table~\ref{tab:navsafe}).  
\ourmethod improves task success and all capability scores across base experts, with larger gains for weaker experts, narrowing the performance gap among them.
This indicates that the runtime compensates for expert-specific failures rather than adding a uniform improvement.
The augmented experts also surpass DrivoR and GTRS-Dense--V2-99 in DS and SR, at the cost of lower comfort, since recovery primitives prioritize admissibility and progress over smoothness.

\begin{table*}[t]
  \centering
  \caption{\textbf{Closed-loop Evaluation on NavSafe-$\infty$.}
  All reported values are three-seed means over the 280-scenario evaluation set.
  Parentheses indicate absolute improvement over the corresponding base method.}
  \label{tab:navsafe}
  \vspace{0.1cm}
  \maintablefont
  \renewcommand{\arraystretch}{1.05}
  \setlength{\tabcolsep}{2pt}
  \begin{tabular*}{\textwidth}{@{\extracolsep{\fill}}>{\raggedright\arraybackslash}p{110pt}
  r@{}l r@{}l
  r@{}l r@{}l
  r@{}l r@{}l
  r@{}l r@{}l@{}}
  \toprule
  \multirow{2}{*}{\textbf{Method}}
  & \multicolumn{8}{c}{\textbf{Closed-Loop Metrics}}
  & \multicolumn{8}{c}{\textbf{Ability Scores}} \\
  \cmidrule(lr){2-9}
  \cmidrule(l){10-17}
  & \multicolumn{2}{c}{DS $\uparrow$}
  & \multicolumn{2}{c}{SR(\%) $\uparrow$}
  & \multicolumn{2}{c}{DE $\uparrow$}
  & \multicolumn{2}{c}{Comf. $\uparrow$}
  & \multicolumn{2}{c}{TC $\uparrow$}
  & \multicolumn{2}{c}{VRUC $\uparrow$}
  & \multicolumn{2}{c}{TV $\uparrow$}
  & \multicolumn{2}{c}{TI $\uparrow$} \\
  \midrule

  DrivoR \citep{kirby2026driving}
  & 64.7 & & 50.7 & & 188.0 & & 87.9 &
  & 58.7 & & 31.7 & & 52.4 & & 43.3 & \\

  GTRS-Dense--V2-99~\citep{Li2025GTRS}
  & 64.3 & & 49.5 & & 163.9 & & 95.9 &
  & 52.0 & & 40.0 & & 54.2 & & 36.7 & \\

  AutoVLA~\citep{zhou2026autovla}
  & 51.1 & & 29.5 & & 186.0 & & 65.4 &
  & 33.0 & & 15.0 & & 29.7 & & 36.7 & \\

  RAP~\citep{feng2026rap}
  & 47.7 & & 27.3 & & \textbf{220.8} & & 48.4 &
  & 32.3 & & 14.2 & & 28.8 & & 22.2 & \\

  MTDrive--SFT~\citep{li2026mtdrive}
  & 48.8 & & 31.8 & & 173.8 & & 94.5 &
  & 30.3 & & 15.0 & & 35.8 & & 44.4 & \\

  \midrule

  LTF \citep{navsim}
  & 53.0 & & 27.7 & & 163.7 & & \textbf{97.8} &
  & 31.7 & & 25.0 & & 28.5 & & 15.6 & \\

  \textbf{+ AutoAgent0}
  \raisebox{-.18\height}{\includegraphics[width=1.4em]{fig/autoagent0_logo.png}}
  & 74.8 & \avggain{21.8} & 60.5 & \avggain{32.8}
  & 160.0 & & 78.8 &
  & 66.0 & \avggain{34.3} & 45.8 & \avggain{20.8}
  & 63.0 & \avggain{34.5} & 52.2 & \avggain{36.6} \\

  \midrule

  DiffusionDrive \citep{diffusiondrive}
  & 44.8 & & 21.0 & & 165.1 & & 97.7 &
  & 21.3 & & 11.7 & & 23.6 & & 22.2 & \\

  \textbf{+ AutoAgent0}
  \raisebox{-.18\height}{\includegraphics[width=1.4em]{fig/autoagent0_logo.png}}
  & 70.9 & \avggain{26.1} & 52.3 & \avggain{31.3}
  & 160.2 & & 72.3 &
  & 52.0 & \avggain{30.7} & 37.5 & \avggain{25.8}
  & 58.8 & \avggain{35.2} & 48.9 & \avggain{26.7} \\

  \midrule

  SparseDriveV2 \citep{sun2026sparsedrivev2}
  & 65.3 & & 48.8 & & 168.0 & & 93.4 &
  & 54.0 & & 43.3 & & 48.2 & & 41.1 & \\

  \textbf{+ AutoAgent0}
  \raisebox{-.18\height}{\includegraphics[width=1.4em]{fig/autoagent0_logo.png}}
  & \textbf{77.7} & \avggain{12.4} & \textbf{63.6} & \avggain{14.8}
  & 165.3 & & 76.0 &
  & \textbf{68.0} & \avggain{14.0} & \textbf{50.0} & \avggain{6.7}
  & \textbf{64.5} & \avggain{16.3} & \textbf{63.3} & \avggain{22.2} \\

  \bottomrule
  \end{tabular*}
\end{table*}

\begin{figure*}[t]
    \centering
    \includegraphics[width=\textwidth]{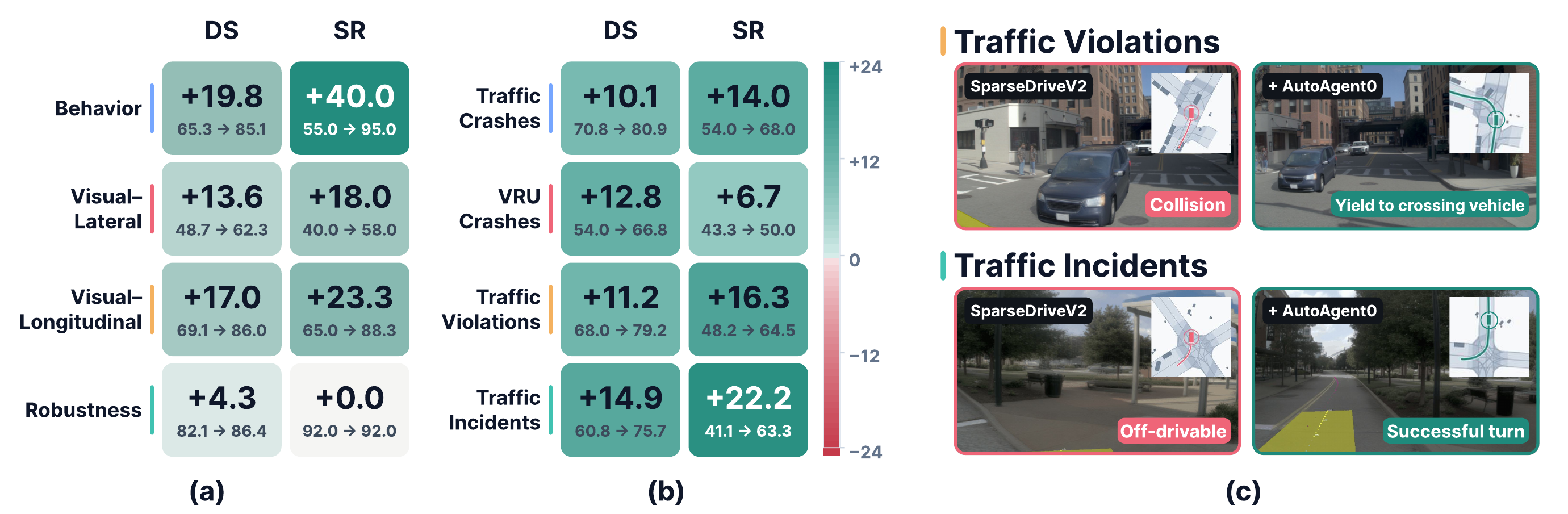}
    \caption{\textbf{Scenario-wise improvements and qualitative examples.}
    DS and SR gains from augmenting SparseDriveV2 with \ourmethod
    on (a) Fail2Drive and (b) NavSafe-$\infty$.
    (c) NavSafe-$\infty$ examples comparing standalone and augmented driving behavior.}
    \label{fig:navsafe_vis}
\end{figure*}

\paragraph{Scenario-wise analysis.}
Figure~\ref{fig:navsafe_vis} shows that the gains of \ourmethod depend on how the expert fails in each scenario group.
Scene context refinement mainly resolves missed pedestrians and animals in Visual--Longitudinal scenarios, while recovery planning drives the largest gains in Behavior scenarios by bypassing or waiting out static blockages.
In Visual--Lateral scenarios, however, many avoided collisions end in stops when no admissible bypass fits the tight geometry, which remains the main open case.
On NavSafe-$\infty$, the largest gains appear in Traffic Incidents, where the examples in (c) show verified yielding and turning (Appendix~\ref{app_sec:scenario_breakdown_and_generalization}).

\vspace{-2mm}

\subsection{Component Analysis and Ablation Studies (RQ2)}
\label{sec:ablations}

\begin{table*}[!htbp]
    \centering
    \caption{\textbf{Ablation of the three runtime modules.}
    HM denotes the harmonic mean of DS and SR.
    Col. is the mean number of collision events per route.
    Rec./km denotes recovery-trigger events per kilometer; standalone SparseDriveV2 has no recovery module, so its Rec./km is zero by construction.
    Routes reaching the 180\,s simulation cap are scored as scenario timeouts (Appendix~\ref{sec:fail2drive-setup}).
    $^\dagger$ Simulator ground-truth scene context replaces the output of scene context refinement.
    $^\ddagger$ Rec./km is not available for this run.}
    \label{tab:recovery-ablation}
    \label{tab:recovery-ablation_1}
    \maintablefont
    \setlength{\tabcolsep}{3pt}
    \renewcommand{\arraystretch}{1.05}
    \begin{tabularx}{\textwidth}{>{\raggedright\arraybackslash}Xccc|ccccc}
        \toprule
        Variant
        & \shortstack{Context\\Refinement}
        & \shortstack{Rule-based\\Verification}
        & \shortstack{Recovery\\Planning}
        & DS $\uparrow$
        & SR (\%) $\uparrow$
        & HM $\uparrow$
        & Col. $\downarrow$
        & Rec./km \\
        \midrule

        SparseDriveV2
        & -- & -- & --
        & 66.5 & 63.4 & 64.7 & 0.445 & 0.0 \\

        + Rule-based verification$^\ddagger$
        & -- & $\checkmark$ & --
        & 69.1 & 60.0 & 64.2 & 0.095 & -- \\

        + Scene context refinement
        & $\checkmark$ & $\checkmark$ & --
        & 76.0 & 74.0 & 75.0 & 0.050 & 54.2\\

        \rowcolor{oursbg}
        \textbf{+ Recovery planning (full)}
        & $\checkmark$ & $\checkmark$ & $\checkmark$
        & 80.0 & 83.4 & 81.7 & 0.145 & 20.3 \\

        \midrule

        \textbf{AutoAgent0$^\dagger$ }
        \raisebox{-.18\height}{%
  \includegraphics[width=1.4em]{fig/autoagent0_logo.png}%
}
        & GT & $\checkmark$ & $\checkmark$
        & 81.4 & 90.3 & 85.6 & 0.110
        & 26.2 \\

        \bottomrule
    \end{tabularx}
\end{table*}

\paragraph{Complementary contributions of runtime components.}
Table~\ref{tab:recovery-ablation} presents cumulative ablations on frozen SparseDriveV2, using a shared controlled fallback when recovery planning is disabled.
Each component addresses the bottleneck left by the previous one.
Verification substantially reduces collisions but slightly lowers task success, since rejecting unsafe proposals without alternatives leaves only conservative fallback.
Scene context refinement then improves both safety and task success, as recovering missed actors prevents collisions and removing spurious detections avoids unnecessary interventions.
Recovery planning further raises task success and sharply reduces trigger frequency, indicating that its maneuvers resolve failures rather than repeatedly contain them, although moving past blockages increases collision events relative to braking (Appendix~\ref{app_sec:recovery_mechanisms}).
Replacing refined context with ground truth yields further gains, identifying perception as the main remaining bottleneck.

\paragraph{Effect of VLM backbone choice.}
We compare Qwen3.5-9B~\citep{qwen2026qwen35}, Opus~5~\citep{anthropic2026opus5}, and GPT-6 Astra~\citep{openai2026gpt6astra} as the VLM backbone within the full \ourmethod runtime on Fail2Drive (Appendix~\ref{app_sec:comparison_of_vlm}).
All three backbones outperform the variant without recovery planning, although Qwen yields lower task success and more collisions.
This gap mainly reflects differences in scene understanding and error correction.
Qwen more frequently hallucinates scene elements and interprets the same scene inconsistently across observations.
Opus and Astra also misjudge motion at times, but they use failure feedback to revise their interpretations and act more conservatively under uncertainty.
Effective recovery therefore depends not only on accurate scene understanding, but also on the ability to correct errors when perception is imperfect.

\vspace{-2mm}

\begin{table*}[t]
    \centering
    \caption{\textbf{Recovery mechanism ablations.}
    Skill denotes access to the primitive library;
    VLM-Choice denotes VLM-based recovery action selection;
    Param denotes parameter selection; and Feedback denotes
    verifier-guided pre-execution revision.
    Budget-matched retry and refinement each allow at most three
    VLM calls per recovery episode.
    Col.\ is the mean number of collision events per route.
    Rec.-SR measures recovery success and is undefined (--)
    for controlled fallback.
    See Appendix~\ref{app_sec:recovery_mechanisms}
    for configurations and evaluation details.}
    \label{tab:recovery-mechanism-ablation}
    \maintablefont
    \setlength{\tabcolsep}{3pt}
    \renewcommand{\arraystretch}{1.05}
    \begin{tabular*}{\textwidth}{@{\extracolsep{\fill}}lcccc|rrrr@{}}
        \toprule
        Variant & Skill & \shortstack{VLM\\Choice} & Param & Feedback
        & DS $\uparrow$ & SR (\%) $\uparrow$
        & Col.\ $\downarrow$ & \shortstack{Rec.-SR\\(\%) $\uparrow$} \\
        \midrule

        Controlled fallback
        & -- & -- & -- & --
        & 76.0 & 74.0 & 0.050 & -- \\

        + Random primitive selection
        & $\checkmark$ & -- & -- & --
        & 74.0 & 73.0 & 0.210 & 52.7 \\

        + VLM skill selection
        & $\checkmark$ & $\checkmark$ & -- & --
        & 76.3 & 77.0 & 0.125 & 61.7 \\

        + Parameterized recovery
        & $\checkmark$ & $\checkmark$ & $\checkmark$ & --
        & 76.3 & 77.3 & 0.200 & 62.5 \\

        + Budget-matched retry
        & $\checkmark$ & $\checkmark$ & $\checkmark$ & --
        & 77.8 & 77.3 & 0.125 & 63.5 \\

        + Verifier-guided refinement
        & $\checkmark$ & $\checkmark$ & $\checkmark$ & $\checkmark$
        & \textbf{80.0} & \textbf{83.4} & 0.145 & \textbf{67.4} \\

        \bottomrule
    \end{tabular*}
\end{table*}

\paragraph{Contribution of recovery mechanisms.}
Table~\ref{tab:recovery-mechanism-ablation} shows that access to recovery primitives alone is insufficient for effective recovery.
Random selection can underperform conservative fallback, suggesting that additional maneuvers are useful only when their selection accounts for the scene.
VLM-based selection improves recovery, whereas parameter selection alone offers limited further gains.
This suggests that refining a maneuver's geometry cannot reliably compensate for choosing an inappropriate maneuver.
Verifier-guided refinement outperforms budget-matched retries, supporting the value of using rejection feedback to reconsider proposals rather than repeatedly generating them from unchanged input.
Its higher collision rate relative to retries also shows that improved task completion does not necessarily imply safer execution.
Together, these results highlight complementary roles for scene-informed maneuver selection and feedback-driven revision, while exposing a remaining safety--progress trade-off.
Appendix~\ref{app_sec:recovery_mechanisms} provides further analysis.


\begin{table*}[t]
    \centering
    \caption{\textbf{Comparison with alternative runtime methods.}
    Reported results cover 80 In-Distribution/Generalization route pairs, excluding calibration routes, with one rollout per route.
    HM is computed within each split; Average is the arithmetic mean of the two splits for each metric.
    Longitudinal MPC preserves the nominal path geometry, while 2D MPC jointly optimizes acceleration and steering.
    AESOP denotes our CARLA adaptation; AESOP + primitives uses our primitive library to provide references for branch MPC.
    Implementation details are provided in
    Appendix~\ref{app:traditional-monitors}.}
    \label{tab:runtime-comparison}
    \maintablefont
    \setlength{\tabcolsep}{4pt}
    \renewcommand{\arraystretch}{1.05}

    \begin{tabularx}{\textwidth}{>{\raggedright\arraybackslash}X ccc ccc ccc}
        \toprule
        \multirow{2}{*}{Method}
        & \multicolumn{3}{c}{In-Distribution}
        & \multicolumn{3}{c}{Generalization}
        & \multicolumn{3}{c}{Average} \\
        \cmidrule(lr){2-4}
        \cmidrule(lr){5-7}
        \cmidrule(lr){8-10}
        & DS $\uparrow$ & SR (\%) $\uparrow$ & HM $\uparrow$
        & DS $\uparrow$ & SR (\%) $\uparrow$ & HM $\uparrow$
        & DS $\uparrow$ & SR (\%) $\uparrow$ & HM $\uparrow$ \\
        \midrule

        SparseDriveV2
        & 70.3 & 75.0 & 72.6
        & 63.1 & 52.5 & 57.3
        & 66.7 & 63.8 & 64.9 \\

        \midrule

        + RSS
        & 73.5 & 66.2 & 69.7
        & 65.5 & 45.0 & 53.3
        & 69.5 & 55.6 & 61.5 \\

        + Longitudinal MPC
        & 69.9 & 55.0 & 61.6
        & 65.4 & 42.5 & 51.5
        & 67.7 & 48.8 & 56.5 \\

        + 2D MPC
        & 65.2 & 65.0 & 65.1
        & 60.1 & 47.5 & 53.1
        & 62.7 & 56.3 & 59.1 \\

        + Uncertainty monitor
        & 74.5 & 60.0 & 66.5
        & 60.7 & 32.5 & 42.3
        & 67.6 & 46.3 & 54.4 \\

        \midrule

        + AESOP (adapted)
        & 56.3 & 52.5 & 54.3
        & 45.7 & 45.0 & 45.3
        & 51.0 & 48.8 & 49.8 \\

        + AESOP + primitives
        & 67.3 & 62.5 & 64.8
        & 57.8 & 60.0 & 58.9
        & 62.6 & 61.3 & 61.8 \\

        \midrule

        \rowcolor{oursbg}
        \textbf{+ AutoAgent0}
        & \textbf{79.8} & \textbf{82.5} & \textbf{81.1}
        & \textbf{81.6} & \textbf{85.0} & \textbf{83.3}
        & \textbf{80.7} & \textbf{83.8} & \textbf{82.2} \\

        \bottomrule
    \end{tabularx}
\end{table*}

\subsection{Comparison with Alternative Runtime Methods (RQ3)}
\label{sec:rq3}

\vspace{-1mm}
\paragraph{Compared runtimes.}
We compare \ourmethod with six alternative runtime configurations around the frozen SparseDriveV2 expert.
RSS requests braking under safe-distance violations~\citep{shalevshwartz2017rss}, longitudinal MPC adjusts speed along the nominal path~\citep{mpcshield}, and the uncertainty monitor requests deceleration based on candidate-score entropy and margins.
The additional 2D MPC baseline jointly optimizes acceleration and steering, allowing lateral adjustments around the expert's nominal trajectory.
Our AESOP adaptation uses anomaly-triggered language-model selection between nominal driving and stop-region recovery options, with branch MPC generating the trajectories.
AESOP + primitives instead supplies references from our primitive library to the same branch MPC, evaluating how these maneuvers contribute within an alternative runtime.
All six baselines replace the nominal trajectory only with verified proposals that achieve a higher PDMS score.

\vspace{-2mm}
\paragraph{Results and analysis.}
Table~\ref{tab:runtime-comparison} shows that \ourmethod achieves the highest DS and SR on both evaluation splits.
Despite its larger action space, 2D MPC achieves a lower DS than longitudinal MPC, although its SR is higher.
This degradation may arise from unsafe lateral maneuvers that introduce additional collisions, offsetting the benefits of greater maneuverability.
The performance gap between our AESOP adaptation and \ourmethod may reflect a mismatch between its intervention pipeline and the challenging scenarios in Fail2Drive, leading to frequent interventions that impede progress.
Nevertheless, incorporating our primitive library improves both DS and SR for the AESOP adaptation, supporting the value of a richer set of recovery maneuvers.

\begin{figure*}[t]
  \centering \includegraphics[width=\textwidth]{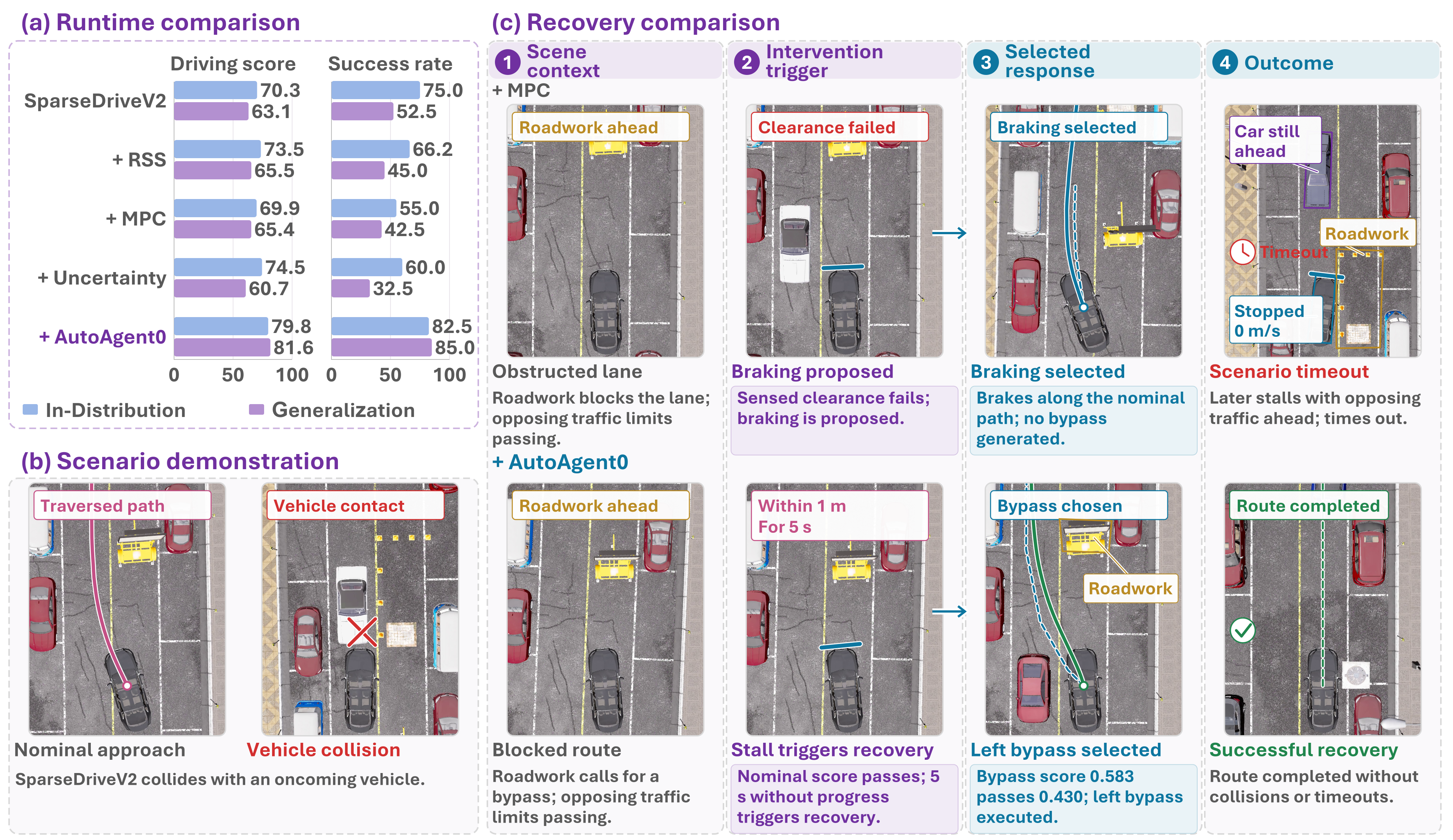}
      \caption{\textbf{Runtime performance and intermediate recovery decisions.}
      \textbf{(a)} DS and SR on 80 In-Distribution/Generalization
      route pairs, excluding calibration routes, with one rollout
      per route.
      \textbf{(b)} SparseDriveV2's maneuver ends in a vehicle collision.
      \textbf{(c)} MPC brakes but later stalls and times out;
      \ourmethod detects stalled progress and executes a checked
      left bypass, completing the route without collisions or timeouts.
      Qualitative rollouts use different configurations;
      see Appendix~\ref{app:traditional-monitors}.}
  \label{fig:traditional-monitor-cases}
\end{figure*}

\paragraph{From intervention to recovery.}
Figure~\ref{fig:traditional-monitor-cases}(b,c) illustrates why intervention and recovery should be distinguished.
Longitudinal MPC brakes but subsequently stalls, whereas \ourmethod detects the lack of progress and executes a verified bypass.
Other traces reveal that a warning may be too transient to trigger intervention, or that a corrective proposal may lose score-based arbitration (Appendix~\ref{app_sec:runtime-comparison-cases}).
Effective recovery therefore depends on the full path from recognizing a failure to selecting, admitting, and executing an alternative that restores progress.
The distinction extends beyond providing lateral controls or a shared primitive library: \ourmethod uses trajectory rejection feedback to revise maneuver selection and search primitive parameters before execution.
The runtime comparison evaluates this integrated design, while the mechanism ablations in Section~\ref{sec:ablations} provide more targeted evidence for the contribution of feedback-guided revision.

%% file: sec/discussion_and_limitations.tex
\vspace{-3mm}
\section{Conclusion}
\label{sec:discussion}

In this paper, we introduce \ourmethod, the first agentic runtime for safe closed-loop driving. Improvements over multiple E2E driving policies across two representative benchmarks suggest that effectiveness of \ourmethod in ensuring safety in safety-critical situations. The comparison with the evaluated runtime monitors further shows the value of using scene context and execution feedback to recover from situations that conservative interventions alone cannot resolve. Importantly, our results highlight \emph{runtime supervision as a complementary direction for improving closed-loop driving beyond advances in the underlying policy}.

\paragraph{Limitations and future works.}
Although \ourmethod improves closed-loop robustness, its verification and recovery remain constrained by scene estimation accuracy and the coverage of the available recovery primitives. While \ourmethod improves task success on NavSafe-$\infty$, the accompanying reductions in comfort and driving efficiency indicate room for improving the quality of recovery behavior. Recovery decisions currently incur second-level VLM latency (Appendix~\ref{app:runtime}), which could be reduced through smaller or distilled models. Future works include: 1) novel discovery of driving skills for the harness and 2) test-time training of E2E policy via \ourmethod's intervention data.


%% file: sec/appendix.tex
\newpage

\section*{%
  \texorpdfstring{%
    \raisebox{-.16\height}{%
      \includegraphics[width=0.06\textwidth]{fig/autoagent0_logo.png}%
    }%
    \hspace{0.5em}%
    \textcolor{myiclrblue}{\textbf{AutoAgent0}}%
    \hspace{0.25em}Appendix%
  }{AutoAgent0 Appendix}%
}

\let\AppendixAddContentsLine\addcontentsline
\renewcommand{\addcontentsline}[3]{%
  \def\RequestedContentsFile{#1}%
  \def\MainContentsFile{toc}%
  \ifx\RequestedContentsFile\MainContentsFile
    \AppendixAddContentsLine{stc}{#2}{#3}%
  \else
    \AppendixAddContentsLine{#1}{#2}{#3}%
  \fi
}

\begingroup
  \hypersetup{linkcolor=black}
  \setcounter{tocdepth}{2}
  \begin{spacing}{0.5}
    \makeatletter
    \@starttoc{stc}
    \makeatother
  \end{spacing}
\endgroup

\clearpage

\input{sec/appendix/a_results_and_analysis}
\input{sec/appendix/b_runtime_efficiency}

\input{sec/appendix/c_runtime_and_scene_refinement}
\input{sec/appendix/d_recovery_planning}
\input{sec/appendix/e_rule_based_verifier}
\input{sec/appendix/f_experimental_protocols}

%% file: sec/appendix/a_results_and_analysis.tex
The appendix provides additional empirical evidence and implementation details for \ourmethod.
Appendix~\ref{app_sec:additional_results} analyzes recovery behavior and performance across models and scenarios;
Appendix~\ref{app:runtime} evaluates computational cost and latency;
Appendices~\ref{app_sec:runtime-implementation}--\ref{app:rules} detail the runtime architecture and scene context refinement, recovery planning, and rule-based verification, respectively;
and Appendix~\ref{app:experimental-details} documents the experimental configurations and evaluation protocols.

\section{Additional Results and Mechanism Analysis}
\label{app_sec:additional_results}
This section supplements the main evaluation of \ourmethod by addressing three questions.
First, how do the recovery ablation variants differ in primitive selection, parameter search, and use of verifier feedback? (Section~\ref{app_sec:recovery_mechanisms})
Second, how do recovery outcomes vary across VLM backbones, and what do execution traces reveal about recovery decisions and monitor interventions? (Sections~\ref{app_sec:recovery_traces},~\ref{app_sec:runtime-comparison-cases}, and~\ref{app_sec:comparison_of_vlm})
Third, where are the gains and remaining limitations concentrated across scenario groups and distribution shifts? (Section~\ref{app_sec:scenario_breakdown_and_generalization})
We provide ablation configurations, qualitative execution traces, VLM backbone comparisons, and scenario-level analyses.

\subsection{Recovery Ablation Configurations}
\label{app_sec:recovery_mechanisms}

Table~\ref{tab:recovery-mechanism-ablation} in Section~\ref{sec:ablations} reports the recovery mechanism ablations.
This section specifies the differences between the evaluated variants.

\paragraph{Controlled fallback.}
Controlled fallback corresponds to the ``+ Scene context refinement'' configuration in Table~\ref{tab:recovery-ablation}.
When no nominal proposal is admitted, it applies conservative braking instead of invoking primitive-based recovery.
Its recovery success rate is undefined because it executes no primitive-based recovery events.

\paragraph{Primitive selection and parameterization.}
Random primitive selection samples uniformly from the primitive menu available at the current decision step.
The menu is filtered by the runtime's availability constraints, rather than sampled from the entire library unconditionally.
This variant removes VLM-based recovery action selection, but does not imply that other runtime modules operate without VLM assistance.
VLM skill selection instead uses the model to choose a primitive, which is instantiated with default parameters.
Parameterized recovery additionally searches the selected primitive's parameter space and verifies the resulting candidate trajectories.
The primitive definitions and trajectory-generation procedure are given in Appendix~\ref{app_sec:programs}.

\paragraph{Repeated generation and feedback-guided revision.}
Budget-matched retry and verifier-guided refinement each permit at most three VLM calls per recovery episode, including the initial call.
The retry variant reuses the initial input without adding previous proposals or their verification outcomes.
The refinement variant incorporates rejection feedback into subsequent selections and excludes the rejected primitive family, allowing the next attempt to consider a different maneuver.
Candidates are verified before execution, and unsuccessful recovery search ends in conservative braking.
The matched call limit controls the number of VLM attempts; the distinguishing factor is whether later attempts receive feedback from earlier ones.

\subsection{Recovery Traces and Failure Cases}
\label{app_sec:recovery_traces}

\begin{figure*}[t]
    \centering
    \includegraphics[width=\textwidth]{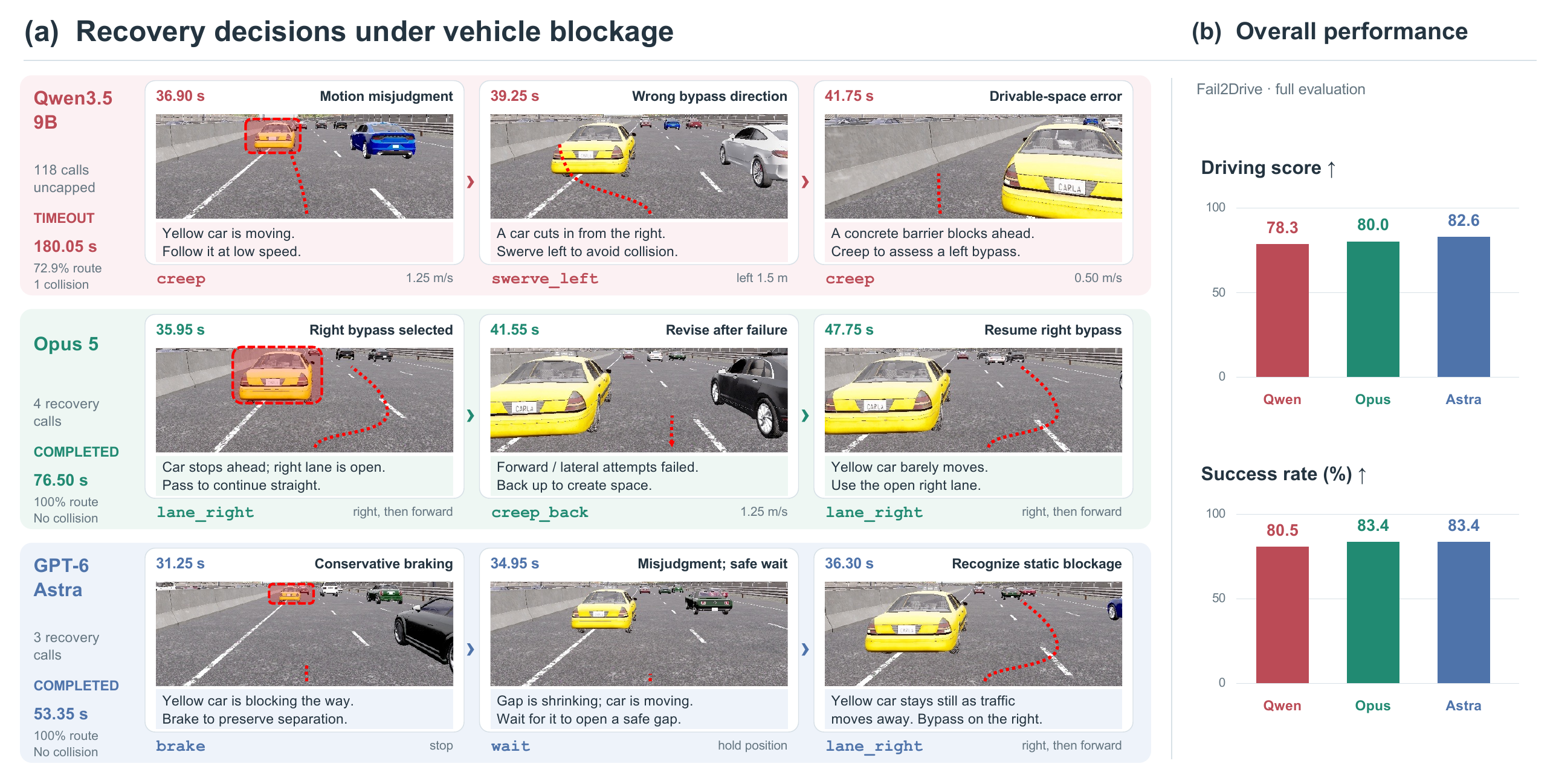}
    \caption{\textbf{Recovery decisions under vehicle blockage.}
    Qwen3.5-9B changes its actions while repeatedly misinterpreting motion and scene constraints, and the rollout ends in timeout with a collision.
    Opus~5 responds to unsuccessful maneuver attempts by reversing before retrying a right bypass.
    GPT-6 Astra initially brakes and waits, then revises its motion interpretation and selects a bypass.
    Both Opus and Astra complete the displayed route without collisions.
    Text annotations summarize the models' stated scene interpretations.
    The right panel reports aggregate benchmark performance.}
    \label{fig:vlm-reasoning-cases}
\end{figure*}

\paragraph{Changing actions does not correct a poor understanding of the scene.}
In Figure~\ref{fig:vlm-reasoning-cases}, Qwen3.5-9B switches from creeping to a left swerve and subsequently returns to creeping, yet the rollout ends with a collision and a timeout at 72.9\% route completion.
During these actions, the model incorrectly interpreted the motion of the blocking vehicle, the correct bypass direction, and the drivable area. This suggests that successive decisions do not establish a reliable account of the constraints preventing progress.
An alternative action can therefore repeat the underlying planning error even when its primitive differs from the previous attempt.
The useful distinction is whether a retry incorporates evidence about why the previous attempt was ineffective and which alternatives remain feasible. Accordingly, retry count and action diversity alone are insufficient measures of recovery quality; the decision sequence must show how execution evidence changes the next proposal.

\paragraph{An intermediate action can make a subsequent maneuver feasible.}
Opus~5 initially selects a right bypass, responds to unsuccessful forward and lateral attempts by reversing, and then retries the right bypass before completing the route without collisions.
This sequence shows that an initial unsuccessful bypass may not be the incorrect maneuver; it may put the vehicle in a more favorable position to perform a reverse.
Although reversing may reduce progress, it allows the model to eventually complete a more favorable bypass.
This implies that recovery primitives should be assessed partly by the conditions they create for subsequent actions, rather than solely by their immediate contribution to route completion.

\paragraph{Conservative actions allow poor scene judgments to be corrected.}
GPT-6 Astra also misjudges the blocking vehicle's motion, but it first brakes and waits before recognizing the persistent blockage and selecting a right bypass.
This shows that recovery capability is determined not only by how accurately a model interprets the scene, but also by which actions it selects under incorrect interpretations and whether those actions give the system an opportunity to revise them.
A temporary motion misjudgment can remain recoverable when the vehicle stays distant from the object, whereas committing to a maneuver that closes that distance under the same uncertainty can expose it to a collision.
However, waiting only helps if the model keeps reassessing the scene; otherwise it stays stopped behind a persistent blockage until it times out.
Good recovery therefore combines cautious actions with a timely switch to another maneuver once the evidence changes.

\subsection{Qualitative Comparison with Traditional Runtime Monitors}
\label{app_sec:runtime-comparison-cases}

\begin{figure*}[t]
  \centering
  \includegraphics[width=\textwidth]{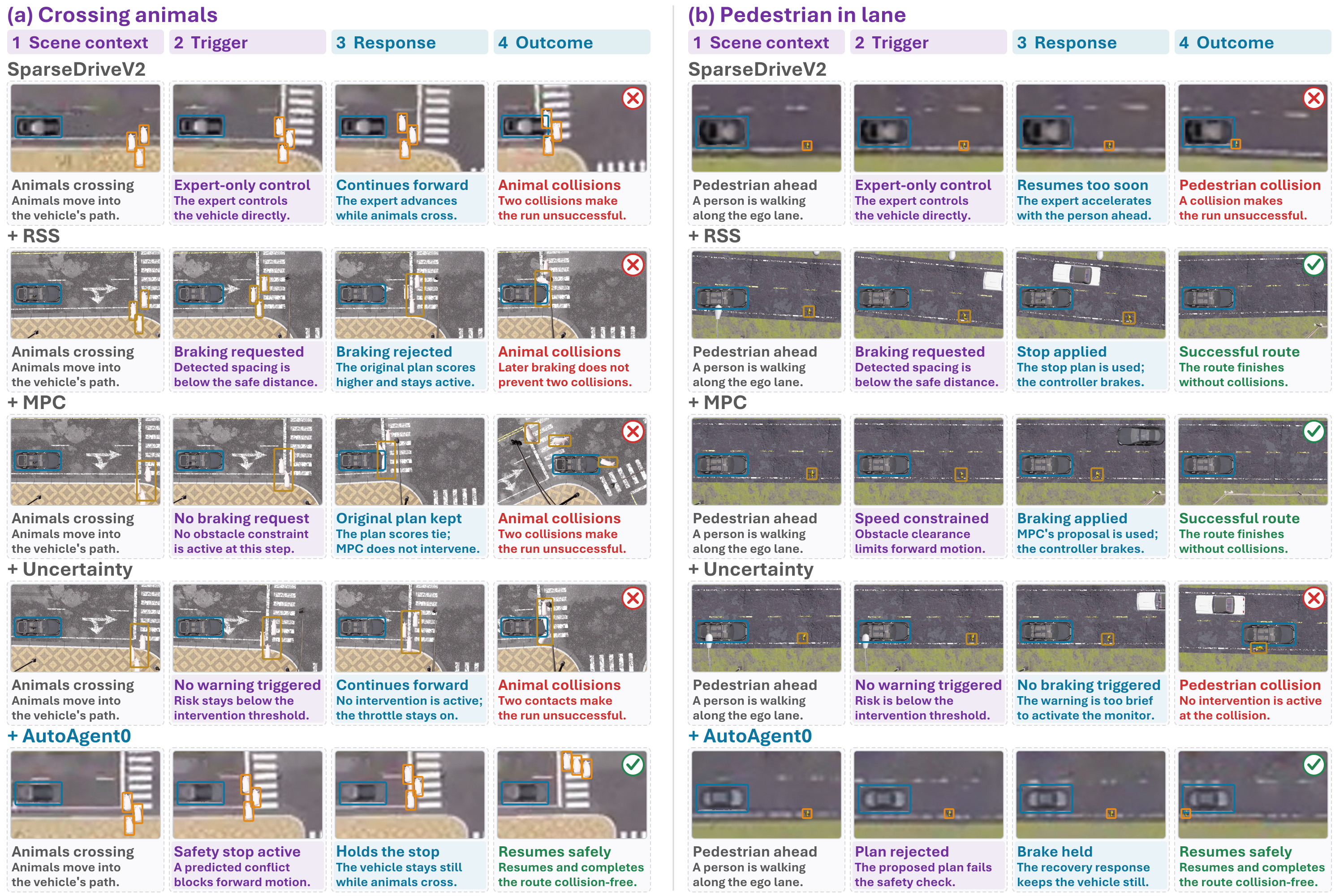}
  \caption{\textbf{Runtime decisions in two safety-critical encounters.}
  Rows compare SparseDriveV2, RSS, MPC, uncertainty monitoring,
  and \ourmethod across scene context, intervention trigger,
  selected response, and outcome.
  \textbf{(a)} Crossing animals: only \ourmethod completes
  without collisions.
  \textbf{(b)} Pedestrian in lane: RSS, MPC, and \ourmethod
  complete without collisions; SparseDriveV2 and uncertainty
  monitoring record a pedestrian collision.
  Blue outlines identify the ego vehicle; gold outlines mark
  hazards.}
  \label{fig:runtime-two-scenes}
\end{figure*}

Figure~\ref{fig:runtime-two-scenes}(a) distinguishes missed interventions from rejected responses. RSS requests braking, but the displayed proposal loses arbitration to the higher-scoring nominal plan; later braking does not prevent collisions. MPC has no active obstacle constraint at the shown step, and uncertainty monitoring does not trigger during the encounter. \ourmethod instead suppresses forward motion and maintains a stop while the animals cross. The distinction is how safety evidence affects execution: detecting a conflict is insufficient if the resulting correction can lose arbitration, whereas enforcing a stop prevents continued motion into the conflict.

In Figure~\ref{fig:runtime-two-scenes}(b), the uncertainty warning does not persist for the three consecutive updates required to trigger intervention, and the vehicle collides with the pedestrian. RSS, MPC, and \ourmethod complete without collisions; \ourmethod rejects a proposed trajectory and holds the brake before resuming. Here, no alternative path is needed, and the failure arises because candidate-score uncertainty does not produce a sustained warning, while \ourmethod's trajectory checks provide a separate basis for rejecting forward motion without waiting for the expert's scores to indicate uncertainty.

\subsection{VLM Backbone Comparison}
\label{app_sec:comparison_of_vlm}

\begin{table*}[t]
        \vspace{0pt}
        \centering
        \caption{\textbf{VLM backbone comparison on Fail2Drive}.
        Rec.-SR denotes recovery success rate.}
        \label{tab:vlm-backbones}
        \small
        \renewcommand{\arraystretch}{1.15}
        \setlength{\tabcolsep}{4pt}{%
            \begin{tabular}{@{}lcccc@{}}
                \toprule
                VLM backbone
                & DS $\uparrow$
                & SR (\%) $\uparrow$
                & Col. $\downarrow$
                & Rec.-SR (\%) $\uparrow$ \\
                \midrule
                Qwen3.5-9B~\citep{qwen2026qwen35} 
                & 78.3 & 80.5
                & 0.210 & 64.6 \\
                Opus 5~\citep{anthropic2026opus5} 
                & 80.0 & 83.4
                & 0.145 & 67.4 \\
                GPT-6 Astra~\citep{openai2026gpt6astra}
                & 82.6 & 83.4
                & 0.125 & 71.0 \\
                \bottomrule
            \end{tabular}%
        }
    \hfill
\end{table*}

\paragraph{Backbone choice affects both task completion and execution quality.}
Table~\ref{tab:vlm-backbones} shows that replacing Qwen3.5-9B with Opus~5 increases DS by 1.7 points and SR by 2.9 percentage points, while reducing Col.\ from 0.210 to 0.145.
The higher completion rate therefore accompanies fewer collisions, indicating that the gain is not simply a consequence of more aggressive execution.
The traces in Section~\ref{app_sec:recovery_traces} illustrate a plausible explanation: revising scene interpretations and unsuccessful maneuvers can restore progress while avoiding repeated commitments to unsuitable actions.
GPT-6 Astra further increases DS by 2.6 points and reduces Col.\ to 0.125, despite matching Opus~5's SR of 83.4\%.
Because SR records binary route outcomes while DS incorporates completion and infraction penalties, this difference shows that a plateau in success rate can conceal improvements in execution quality.
Although GPT-6 Astra achieves the highest DS, we use Opus~5 as the default backbone throughout the paper so that all ablations and comparisons share a single VLM configuration and remain directly comparable.

\paragraph{Local recovery gains must be assessed through subsequent route outcomes.}
Astra improves Rec.-SR over Opus~5 from 67.4\% to 71.0\% without increasing aggregate route success.
The metrics evaluate different stages of execution: a recovery event ends after a sustained return to nominal driving, whereas route success requires completing the remaining task.
A locally successful recovery can consequently precede another blockage, collision, or timeout; different backbones may also encounter different recovery-event populations.
Thus, a higher Rec.-SR does not directly establish that more routes were rescued, and equal SR does not imply that the same routes succeeded.
This distinction suggests a concrete way to assess recovery quality: track the final outcomes of routes that recover, and distinguish recurrence of the original blockage from failures caused by subsequent hazards.
Such an analysis would reveal whether a backbone merely restores nominal execution temporarily or places the vehicle in a state from which the task can be completed reliably.

\subsection{Scenario Breakdown and Generalization}
\label{app_sec:scenario_breakdown_and_generalization}

\begin{figure*}[t]
    \centering
    \includegraphics[width=\textwidth]{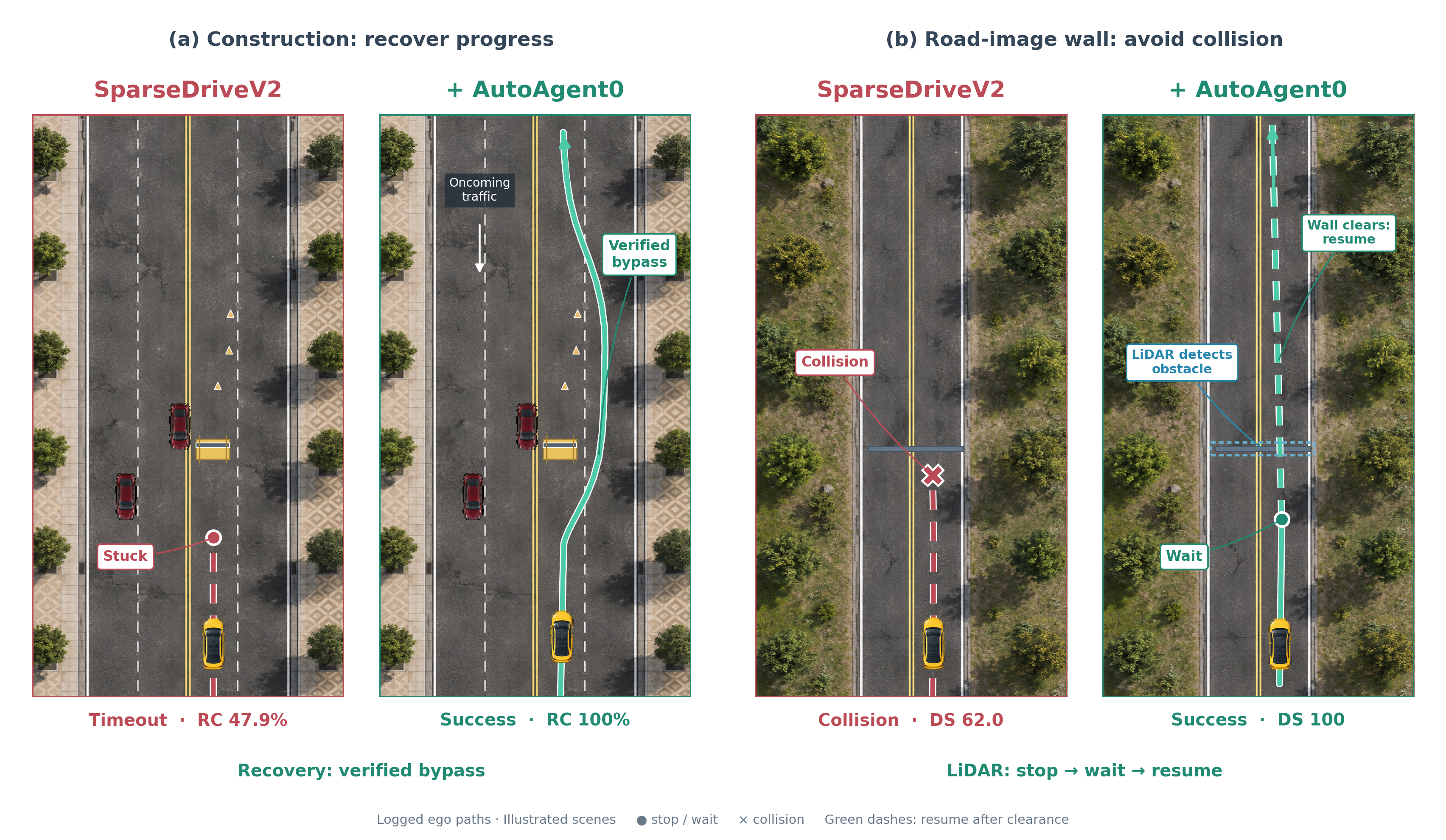}
    \caption{\textbf{Qualitative examples of complementary runtime behaviors.}
    Left: a verified bypass resolves a construction blockage.
    Right: obstacle detection supports stopping before a road-image wall and resuming after it clears.
    The examples illustrate improvements through both active maneuvering and appropriate waiting.}
    \label{fig:behavior-recovery-cases}
\end{figure*}

\paragraph{Gains are concentrated in scenarios that require behavioral adaptation.}
Figure~\ref{fig:navsafe_vis}(a) shows the largest SR improvement in the Behavior group: DS increases by 19.8 points and SR by 40.0 percentage points.
Thus, the increase in successful routes is much larger than the change in DS.
This pattern is consistent with resolving failures that averages can obscure, including blockages after substantial progress and unsafe executions that still traverse much of the route.
The Visual--Longitudinal group exhibits a related pattern: DS increases by 17.0 points and SR by 23.3 percentage points.
Here too, the benefit is expressed primarily in successful and penalty aware execution.

\paragraph{Improved success does not imply uniform improvement.}
For Visual--Lateral scenarios, DS and SR increase by 13.6 points and 18.0 percentage points, respectively, but both remain the lowest of the four groups.
The Robustness group has a smaller DS increase of 4.3 points and unchanged aggregate SR.
The current results therefore do not show an increase in successful routes for every scenario group.
The runtime's benefits depend on the type of failure encountered, and some conditions remain outside the capabilities demonstrated by the present recovery mechanisms.

\paragraph{Recovery includes deciding when to move and when to wait.}
The two Behavior examples in Figure~\ref{fig:behavior-recovery-cases} illustrate complementary responses.
In the construction case, the standalone expert stalls at 47.9\% completion, whereas a verified bypass completes the route.
In the road-image wall case, the standalone execution collides; the augmented system detects the obstacle, waits, and resumes after clearance.
The first case requires an alternative path, while the second requires withholding motion until the environment permits progress.
A recovery policy that treats all low-progress states as a reason to maneuver would fail to distinguish these situations.
Scene context determines whether progress should be restored through action or through waiting.

\paragraph{Generalization gains exceed in-distribution gains.}
Using the split-level values in Table~\ref{tab:main-results}, the DS generalization gap, $\mathrm{DS}_{\mathrm{ID}}-\mathrm{DS}_{\mathrm{Gen}}$, changes from 3.5 to $-0.8$ for UniAD, from 7.1 to 1.5 for SparseDriveV2, and from 9.7 to 6.0 for BridgeDrive.
The corresponding DS gains on in-distribution versus generalization routes are 20.9 versus 25.2 points for UniAD, 10.7 versus 16.3 points for SparseDriveV2, and 0.1 versus 3.8 points for BridgeDrive.
These differences are computed from the rounded values reported in the table.

The gap reductions therefore accompany improvements on both splits, rather than degradation of in-distribution performance.
BridgeDrive is particularly informative: its already high in-distribution performance changes little, while its generalization SR increases from 75.0\% to 85.3\%.
This suggests that runtime augmentation can address residual long-tail failures even when ordinary-route performance leaves limited room for improvement.

%% file: sec/appendix/b_runtime_efficiency.tex
\section{Runtime Efficiency and Latency}
\label{app:runtime}

\subsection{Module-level Latency}

We profile the execution time of individual modules to characterize the computational cost of \ourmethod.
Table~\ref{tab:module_latency} reports local module latencies measured during closed-loop driving.
Nominal planning and perception inference have median latencies of 397.4 and 303.9\,ms, respectively.
Batch trajectory verification takes 78.6\,ms, while single-candidate verification takes 12.0\,ms.
Recovery temporal search takes 468.6\,ms per invocation, and the tracking controller takes 3.3\,ms.
The contribution of these operations to overall runtime also depends on their invocation frequency, which we examine in the next subsection.

\begin{table}[t]
\centering
\small
\caption{Local module wall-clock latency across 12 profiling routes.
Statistics pool completed calls after excluding the first 20 control steps of each route.
Timings include inter-process communication and nested local operations, so rows are not additive.
VLM request latency is measured separately in Table~\ref{tab:vlm_request_latency}.}
\label{tab:module_latency}
\setlength{\tabcolsep}{5pt}
\begin{tabular}{lrrrr}
\toprule
Module & Calls & Mean (ms) & Median (ms) & P95 (ms) \\
\midrule
Nominal planning
    & 9,227 & 598.9 & 397.4 & 1,326.7 \\
Perception inference
    & 3,017 & 460.5 & 303.9 & 939.4 \\
Perception update without inference
    & 9,032 & 112.9 & 106.2 & 169.1 \\
Perception fusion and refinement
    & 3,017 & 178.2 & 169.7 & 344.3 \\
Scene-context resolution
    & 12,049 & 0.21 & 0.19 & 0.37 \\
Batch trajectory verification
    & 9,227 & 80.2 & 78.6 & 153.1 \\
Single-candidate verification
    & 15,439 & 13.9 & 12.0 & 31.4 \\
Recovery temporal search
    & 68 & 503.5 & 468.6 & 1,100.2 \\
Tracking controller
    & 12,049 & 3.28 & 3.27 & 6.07 \\
Post-step visualization and logging
    & 12,049 & 82.7 & 66.4 & 197.3 \\
\bottomrule
\end{tabular}
\end{table}

\paragraph{Profiling protocol.}
We use SparseDriveV2 as the nominal planner, the same adapted BridgeDrive detector used in the closed-loop evaluations as the perception module, and Opus~5 as the VLM backend.
The profiling set contains 12 Fail2Drive routes, comprising one in-distribution route and one generalization route from each of six scenario types.
Each route is evaluated once, yielding 12,289 control steps in total.
Excluding the first 20 steps of each route leaves 12,049 steps for the reported statistics.
We measure elapsed wall-clock time at module boundaries, including inter-process communication, and record initialization separately.
The experiments use shared NVIDIA L40S GPUs, with at most one profiling episode per GPU at a time.
The reported latencies therefore characterize the observed shared-resource conditions.

\paragraph{VLM request latency.}
We separately measure VLM request latency by replaying the same 20 saved recovery-decision contexts through Qwen3.5-9B, GPT-6 Astra, and Opus~5.
The contexts are drawn from 11 of the profiling routes.
Each context preserves the original task prompt, system prompt, and two front-camera images, ordered from the historical observation to the current observation.
Each model processes all 20 contexts serially, without running CARLA or the driving pipeline.
Model loading is excluded, while the first request is retained; no additional warm-up requests or automatic retries are used.
We time the complete request call, including input preparation and response parsing.
For API backends, this measurement also includes network and service latency.

Table~\ref{tab:vlm_request_latency} summarizes the results.
Median request latency is 8.89\,s for Qwen3.5-9B, 4.36\,s for GPT-6 Astra, and 4.69\,s for Opus~5.
All 60 requests return parseable JSON without reported request errors or output truncation.
These results characterize recovery-decision requests; stop-sign and semantic-takeover checks are not included in this replay experiment.

\begin{table}[t]
\centering
\small
\caption{Standalone VLM request latency on the same 20 saved recovery contexts.
Each model processes one request per context.
Qwen runs locally on a shared L40S, while Astra and Opus use remote APIs.
The measurements exclude model loading but include input preparation and response parsing.}
\label{tab:vlm_request_latency}
\setlength{\tabcolsep}{7pt}
\begin{tabular}{lrrrr}
\toprule
Model & Requests & Mean (s) & Median (s) & P95 (s) \\
\midrule
Qwen3.5-9B  & 20 & 8.33 & 8.89 & 9.68 \\
GPT-6 Astra & 20 & 4.72 & 4.36 & 6.82 \\
Opus 5     & 20 & 4.77 & 4.69 & 6.39 \\
\bottomrule
\end{tabular}
\end{table}

\paragraph{VLM configuration and measurement scope.}
Qwen3.5-9B uses BF16 weights, greedy decoding, and disabled thinking, without quantization or CPU offloading.
Its attention implementation uses SDPA, while the DeltaNet layers use the PyTorch implementation because the optional accelerated kernels are unavailable in the measurement environment.
Opus~5 uses low effort with thinking disabled, while GPT-6 Astra uses low reasoning effort.
The output allowance is 800 tokens; the Astra adapter adds a 1,024-token reasoning reserve, giving a combined output cap of 1,824 tokens.

\subsection{Invocation Frequency and Waiting Behavior}

Per-call latency alone does not determine the runtime cost of a module.
We consider invocation frequency, execution behavior while results are pending, and the age of the observations supporting an applied decision.

\paragraph{Invocation frequency.}
Over the 12,049 control steps retained for the local latency analysis, nominal planning and batch verification are each invoked 9,227 times.
Perception inference is invoked 3,017 times, while the remaining 9,032 steps use perception updates without a new inference call.
Recovery temporal search is invoked 68 times across 11 of the 12 routes, corresponding to approximately 0.56 calls per 100 control steps.
Thus, although recovery search has a substantial per-call cost, it is invoked much less frequently than nominal planning or perception inference in this sample.
Single-candidate verification is invoked 15,439 times, reflecting that a control step may contain multiple verification calls.

The full closed-loop traces, including the initial 20 steps of each route, contain 60 completed recovery VLM calls, 17 completed stop-sign checks, and seven completed semantic-takeover checks.
These counts refer to Opus~5 calls in the closed-loop experiment, rather than the standalone replay requests in Table~\ref{tab:vlm_request_latency}.

\paragraph{Waiting behavior.}
In the synchronous Fail2Drive execution path, recovery VLM requests block the current agent step.
The evaluator does not advance the simulation while a recovery request is pending.
Thus, a several-second wall-clock request does not correspond to the same duration of simulated driving or braking.
After the request returns, recovery candidates are verified using the scene context of the same agent step, and the returned control is applied at the next simulation tick.
Object-level semantic review follows a separate asynchronous path, with returned decisions checked for recency and track consistency before use.
In the closed-loop Opus~5 traces, the median backend round-trip time is 4.18\,s for recovery requests and 2.95\,s for pooled stop-sign and semantic-takeover requests.

\paragraph{Real-time applicability.}
Deterministic trajectory checks and execution guards do not require a fresh VLM decision at every step, but this separation does not establish real-time emergency response.
Because synchronous recovery pauses the simulator, the reported safety gains do not demonstrate protection during an equivalent real-world reasoning delay.
Deployment would require explicit handling of stale observations and pending decisions, together with independently running low-level safety control.

\subsection{Route-level Efficiency}
We measure the closed-loop cost of adding \ourmethod with the Qwen3.5-9B VLM as the backbone to SparseDriveV2 on
the same 200 Fail2Drive routes. \ourmethod increases the median wall-clock
time per route from $9.88$ to $11.44$\,min ($+15.7\%$), while reducing the
median simulated time from $44.4$ to $38.9$\,s ($-12.5\%$).
Table~\ref{tab:runtime} summarizes both quantities.

\paragraph{Protocol.}
Both evaluations use identical route IDs, and the recorded route lengths
match for all 200 routes. Wall-clock time spans the first to the last agent
step of the final completed attempt, excluding world loading and agent
setup. Simulated time is the CARLA game time at the last step. The
\ourmethod evaluation enforces a 180\,s simulation limit, whereas the
baseline evaluation did not, and 19 baseline routes ran for up to 372\,s.
We truncate those routes at 180\,s of simulated time and use the wall-clock
time at that step.

\paragraph{Results.}
\ourmethod takes longer in wall-clock time on 132 of 200 routes, with a
median per-route ratio of $1.17\times$. The mean increases by only $2.6\%$
because a few slow baseline routes dominate the baseline mean. Simulating one
second of driving takes $16.61$\,s of wall-clock time with \ourmethod,
compared with $14.85$\,s for the baseline ($+11.8\%$ pooled). Shorter
episodes partly offset this per-step overhead: \ourmethod finishes in less
simulated time on 148 routes, and nine routes tie at the 180\,s limit.
\begin{table}[t]
\centering
\small
\caption{Per-route runtime on the same 200 Fail2Drive routes under a matched
180\,s simulation budget. The \ourmethod in this experiment uses the Qwen3.5-9B VLM as its backbone. Wall-clock time is in minutes and simulated time
in seconds. Wall-clock per simulated second is the real time needed to
simulate one second of driving; the pooled rate divides total wall-clock by
total simulated time. $\Delta$ is the relative change of \ourmethod in
percent.}
\label{tab:runtime}
\setlength{\tabcolsep}{5pt}
\begin{tabular}{llrrr}
\toprule
Quantity & Statistic & SparseDriveV2 & + AutoAgent0 & $\Delta$ \\
\midrule
Wall-clock (min) & Mean      & 15.49 & 15.90 & $+2.6$ \\
                 & Median    & 9.88  & 11.44 & $+15.7$ \\
                 & Total (h) & 51.6  & 53.0  & $+2.6$ \\
Simulated (s)    & Mean      & 62.6  & 57.4  & $-8.2$ \\
                 & Median    & 44.4  & 38.9  & $-12.5$ \\
                 & Routes at 180\,s & 19 & 21 & -- \\
Wall-clock per   & Pooled    & 14.85 & 16.61 & $+11.8$ \\
simulated second (s) & Median & 12.13 & 17.17 & $+41.5$ \\
\bottomrule
\end{tabular}
\end{table}

%% file: sec/appendix/c_runtime_and_scene_refinement.tex
\section{Runtime Architecture and Scene Context Refinement}
\label{app_sec:runtime-implementation}

\subsection{Runtime Control Flow}
\label{app_sec:runtime-control-flow}

Algorithm~\ref{alg:harness} summarizes the closed-loop runtime.
At each step, the runtime updates its memory with execution feedback and refines the scene context.
When a new planning decision is needed, nominal proposals are checked first; recovery search is invoked only if all nominal candidates are rejected.
The search uses verifier feedback to revise recovery proposals within a bounded budget, committing a proposal only after it passes verification.
An ongoing recovery maneuver is advanced under execution guards before a new planning decision is made.
If no trajectory is available, the runtime applies conservative braking.

\begin{algorithm}[htbp]
\caption{\ourmethod Runtime with Scene Context Refinement and Agentic Recovery Planning}
\label{alg:harness}
\small
\begin{algorithmic}[1]
\Require Nominal policy $\pi_{\mathrm N}$, route $q$,
    scene-refinement skills $\mathcal F$, action primitive library $\mathcal P$,
    rule-based verifier $V$, search budget $B$
\State Initialize $\mathit{memory} \gets \varnothing$,
    $\mathit{recovery\_state} \gets \mathrm{None}$, and $t \gets 0$

\While{the route is active}
    \State Receive observation $o_t$ and update history $h_t$
    \State Record execution feedback from $o_t$ and $\mathit{recovery\_state}$ in $\mathit{memory}$
    \State $c_t \gets \Call{BuildContext}{h_t,q,\mathit{memory}}$

    \Statex \textbf{Scene Context Refinement}
    \State $\widetilde c_t \gets
        \Call{SceneContextRefinement}{\mathcal F,c_t}$

    \State $\tau_t \gets \mathrm{None}$
    \If{$\mathit{recovery\_state} \neq \mathrm{None}$}
        \State Advance committed recovery using $\widetilde c_t$ and execution guards;
        \Statex \hspace{2\algorithmicindent} update $\tau_t$, $\mathit{recovery\_state}$, and $\mathit{memory}$
    \EndIf

    \If{$\tau_t = \mathrm{None}$}
        \Comment{A new planning decision is needed}
        \State $\mathit{candidates} \gets
            \Call{ProposeNominal}{\pi_{\mathrm N},c_t}$

        \Statex \hspace{\algorithmicindent}
            \textbf{Rule-based Verification}
        \State $(\mathit{accepted},\mathit{feedback}) \gets
            \Call{VerifyCandidates}{V,\mathit{candidates},\widetilde c_t}$

        \If{$\mathit{accepted} \neq \varnothing$}
            \State $\tau_t \gets
                \Call{SelectBest}{\mathit{accepted}}$
                \Comment{Verified nominal policy $\pi_{\mathrm N}^{V}$}
        \Else \Comment{All nominal candidates were rejected}
            \Statex \hspace{2\algorithmicindent}
                \textbf{Recovery Planning via Agentic Parameter Search}
            \State Record $\mathit{candidates}$ and $\mathit{feedback}$ in $\mathit{memory}$

            \For{$j=1,\ldots,B$}
                \State $(\ell,\Theta) \gets
                    \Call{SelectAndRefine}{\mathcal P,\widetilde c_t,\mathit{feedback},\mathit{memory}}$
                \State $\mathit{candidates} \gets
                    \Call{GenerateCandidates}{\ell,\Theta,\widetilde c_t}$
                \State $(\mathit{accepted},\mathit{feedback}) \gets
                    \Call{VerifyCandidates}{V,\mathit{candidates},\widetilde c_t}$
                \State Record $\ell$, $\Theta$, $\mathit{candidates}$, and $\mathit{feedback}$ in $\mathit{memory}$

                \If{$\mathit{accepted} \neq \varnothing$}
                    \State $\tau_t \gets
                        \Call{SelectBest}{\mathit{accepted}}$
                    \State $\mathit{recovery\_state} \gets
                        \Call{CommitRecovery}{\ell,\tau_t,\widetilde c_t}$
                    \State \textbf{break}
                \EndIf
            \EndFor
        \EndIf
    \EndIf

    \If{$\tau_t = \mathrm{None}$}
        \State $\tau_t \gets \Call{ConservativeBraking}{o_t}$
            \Comment{No proposal admitted}
    \EndIf

    \State \Call{ExecuteSegment}{$\tau_t$}
        \Comment{Tracking controller; feedback arrives in $o_{t+1}$}
    \State $t \gets t+1$
\EndWhile
\end{algorithmic}
\end{algorithm}

Here, $\ell$ denotes a recovery primitive and $\Theta$ a set of parameter
vectors within its allowed domain. \textsc{GenerateCandidates} applies the
primitive's trajectory generator to each parameter vector using the refined
context $\widetilde c_t$. The variables \textit{candidates} and
\textit{accepted} are reused for nominal and recovery proposals, while
\textit{feedback} records the verifier's assessment.

\subsection{Scene Context Refinement}
\label{app:scene-context-refinement}

\paragraph{Geometric Refinement and Disagreement Detection.}
The perception pipeline supports bounded refinement of predicted object boxes using associated LiDAR and radar returns.
LiDAR refinement uses nearby non-ground points, while radar refinement applies to eligible forward vehicle and static-object boxes.
An optional disagreement module compares the original detector boxes with LiDAR evidence from the same sensor capture.
It identifies two types of candidates: detected objects with insufficient LiDAR support, and non-ground LiDAR clusters not explained by current detections.
These disagreements identify objects for further inspection rather than directly establishing false positives or false negatives.
Invalid LiDAR evidence does not count as evidence that an object is absent.

\paragraph{Optional Object-level Semantic Validation.}
When enabled, the disagreement module associates candidates across observations using their positions and dimensions, and accumulates disagreement observations within a sliding time window.
Candidates become eligible for VLM review after reaching a configured observation count, subject to request cooldowns and backend availability.
The VLM receives a camera image with the candidate's projected location marked, an unmarked crop of the same region, and a description of the sensor disagreement.
It returns a \texttt{keep}, \texttt{drop}, or \texttt{uncertain} decision together with a confidence value and a short explanation.
Requests run asynchronously, allowing the control loop to continue while a review is pending.

\paragraph{Applying Validated Corrections.}
A returned decision is accepted only if it satisfies the confidence threshold, remains sufficiently recent, and still matches the active candidate.
An accepted \texttt{keep} decision retains a matching object or adds the candidate to the obstacle list if no match is present.
An accepted \texttt{drop} decision removes matching objects from that list.
Uncertain, malformed, or expired responses do not introduce a new correction.
Cached decisions expire after a configured lifetime, and renewed agreement between the sensors invalidates an earlier decision for the corresponding track.
The resulting object list supplies the corrected obstacle evidence used in subsequent trajectory verification.

%% file: sec/appendix/d_recovery_planning.tex
\section{Agentic Recovery Planning}
\label{app_sec:programs}

\subsection{Primitive Library and Trajectory Generation}
\label{app_sec:recovery-primitives}

Table~\ref{tab:primitives} summarizes the fifteen primitives in the
recovery library. Each primitive defines a maneuver that the
deterministic runtime converts into a candidate trajectory.
The table reports default durations and parameter ranges;
the available actions and effective parameter ranges may be
restricted by the runtime configuration and current scene.
For example, the runtime excludes \texttt{swerve\_left} and
\texttt{swerve\_right} when the ego speed is below 1\,m/s.

The VLM selects from the available primitives, while the runtime
handles trajectory generation and verification. Parameter search
is described in Section~\ref{app_sec:recovery-parameter-search},
and execution-time safety checks are described in
Section~\ref{app_sec:recovery-execution}.

\begin{table}[htbp]
    \centering
    \caption{\textbf{Recovery primitive library.}
    Durations and parameter ranges describe the default action
    specifications. Actual execution may end early or use a
    configuration-dependent duration. A dash indicates that no
    parameter range is listed in the default specification.}
    \label{tab:primitives}
    \small
    \renewcommand{\arraystretch}{1.12}
    \setlength{\tabcolsep}{4pt}
    \resizebox{\textwidth}{!}{%
    \begin{tabular}{@{}l>{\raggedright\arraybackslash}p{0.43\textwidth}cl@{}}
        \toprule
        Primitive & Behavior & Default duration & Parameter range \\
        \midrule
        brake
        & Brake to a full stop
        & 1.0\,s & -- \\

        wait
        & Remain stopped
        & 1.0\,s & -- \\

        creep
        & Crawl straight forward at low speed
        & 1.5\,s & speed 0.5--2.0\,m/s \\

        creep\_back
        & Stop, then back straight up with heading preserved
        & 3.0\,s & speed 0.5--2.0\,m/s \\

        swerve\_left
        & Swerve left around an obstacle, then stop
        & 2.5\,s & lateral 0.5--2.5\,m \\

        swerve\_right
        & Swerve right around an obstacle, then stop
        & 2.5\,s & lateral 0.5--2.5\,m \\

        back\_left
        & Stop, reverse while steering left, then stop
        & 5.0\,s & \begin{tabular}[t]{@{}l@{}}
            lateral 0.3--1.5\,m \\
            speed 0.5--2.0\,m/s
        \end{tabular} \\

        back\_right
        & Stop, reverse while steering right, then stop
        & 5.0\,s & \begin{tabular}[t]{@{}l@{}}
            lateral 0.3--1.5\,m \\
            speed 0.5--2.0\,m/s
        \end{tabular} \\

        back\_bypass\_left
        & Reverse, then drive forward around the obstacle on the left
        & 6.0\,s & \begin{tabular}[t]{@{}l@{}}
            lateral 1.0--3.0\,m \\
            back speed 0.5--3.0\,m/s \\
            forward speed 0.5--4.0\,m/s
        \end{tabular} \\

        back\_bypass\_right
        & Reverse, then drive forward around the obstacle on the right
        & 6.0\,s & \begin{tabular}[t]{@{}l@{}}
            lateral 1.0--3.0\,m \\
            back speed 0.5--3.0\,m/s \\
            forward speed 0.5--4.0\,m/s
        \end{tabular} \\

        three\_point
        & Perform a bounded three-point turn
        & 6.0\,s & \begin{tabular}[t]{@{}l@{}}
            forward speed 0.5--4.0\,m/s \\
            back speed 0.5--2.0\,m/s
        \end{tabular} \\

        lane\_left
        & Change one lane to the left and finish aligned
        & 3.0\,s & \begin{tabular}[t]{@{}l@{}}
            lateral 1.0--3.5\,m \\
            speed 0.5--2.0\,m/s
        \end{tabular} \\

        lane\_right
        & Change one lane to the right and finish aligned
        & 3.0\,s & \begin{tabular}[t]{@{}l@{}}
            lateral 1.0--3.5\,m \\
            speed 0.5--2.0\,m/s
        \end{tabular} \\

        lane\_bypass\_left
        & Borrow the adjacent left lane, pass a static obstacle,
          and return to the original lane
        & 6.0\,s & \begin{tabular}[t]{@{}l@{}}
            longitudinal 3.0--20.0\,m \\
            lateral 0.4--4.0\,m \\
            speed 6.0--18.0\,m/s
        \end{tabular} \\

        lane\_bypass\_right
        & Borrow the adjacent right lane, pass a static obstacle,
          and return to the original lane
        & 6.0\,s & \begin{tabular}[t]{@{}l@{}}
            longitudinal 3.0--20.0\,m \\
            lateral 0.4--4.0\,m \\
            speed 6.0--18.0\,m/s
        \end{tabular} \\
        \bottomrule
    \end{tabular}%
    }
\end{table}

\subsection{Parameter Search and Verifier-Guided Refinement}
\label{app_sec:recovery-parameter-search}

Recovery planning starts with the VLM selecting a primitive based on the
scene context and the reason for nominal-plan rejection. The runtime then
samples parameters within the selected primitive's allowed range and
generates candidate trajectories. Each candidate is evaluated by the
rule-based verifier described in Appendix~\ref{app:rules}, and the
highest-scoring accepted candidate is selected for execution.

If no candidate passes verification, the rejection feedback is returned
to the VLM to guide another primitive selection. This process continues
within a bounded search budget. If no acceptable trajectory is found,
the runtime falls back to conservative braking.

\subsection{VLM Prompts and Invocation Protocol}
\label{app_sec:recovery-vlm-protocol}

The VLM receives front-camera observations, the route instruction, scene
context, and verifier feedback, together with the available recovery
primitives. The prompt asks it to identify the blockage, determine a
suitable escape direction, and select a recovery primitive. When earlier
attempts have failed, their outcomes are included to inform the next
decision.

The response specifies a primitive and a brief explanation. The runtime
checks that the selected primitive belongs to the available set before
performing parameter search. Numerical trajectory generation and
verification are handled by the runtime. Invalid responses or exhausted
invocation budgets trigger the fallback procedure.

\paragraph{Prompt format.}
We provide a condensed template of the recovery-plan prompt below.
Scene-dependent fields are filled at runtime, and front-camera images
are supplied as visual inputs. The available action list is filtered
by the runtime before invocation. The VLM identifies the blockage,
selects an escape direction, and proposes candidate primitives for
subsequent sampling and verification.

\begin{quote}
\small
\begin{verbatim}
[System]
You are a driving recovery planner.
Use only exact primitive IDs supplied in the user prompt.
Return only valid JSON.

[Visual input]
<Front-camera image(s)>

[Driving context]
Route direction: <route instruction>
Road context: <lane and road information>

[Previous recovery result]
<Previous action and its outcome>

[Recovery memory]
<Relevant earlier attempts and feedback>

[Safe primitive candidates]
<Available primitive IDs, descriptions, and speed ranges>

[Task]
1. Describe the visible blockage in one short sentence.
2. Choose an escape direction:
   forward, left, right, back, or stop.
3. Return up to five candidate primitives with probabilities
   and brief reasons.
4. Use only the supplied primitive IDs.
5. Specify target_speed_mps only when supported;
   otherwise use null.
6. If no candidate is appropriate, choose stop and
   return an empty candidates list.

[Output]
Return JSON only, without Markdown fences.
\end{verbatim}
\end{quote}

\paragraph{Response format.}
The following example illustrates the response structure.
The action must belong to the supplied candidate list.

\begin{quote}
\small
\begin{verbatim}
{
  "front_blockage": "A static obstacle blocks the route lane.",
  "escape_direction": "left",
  "candidates": [
    {
      "primitive": {
        "action": "lane_left",
        "target_speed_mps": null
      },
      "probability": 1.0,
      "reasoning": "The adjacent left lane provides space."
    }
  ]
}
\end{verbatim}
\end{quote}

The runtime validates the response and uses the candidate probabilities
to allocate the sampling budget. Generated trajectories are verified
before execution. When another selection is requested, feedback from
rejected attempts is included in the next prompt.

\subsection{Execution, Interruption, and Fallback}
\label{app_sec:recovery-execution}

Once a recovery trajectory passes verification, the runtime commits it
to the tracking controller. The trajectory is anchored in the world
frame and advanced as the vehicle moves, allowing multi-stage maneuvers
to proceed without restarting at each control step.

During execution, the runtime monitors progress and applies
maneuver-specific safety guards using updated scene observations.
These execution guards are distinct from the full admission check used
during candidate selection. A detected hazard or stalled maneuver can
interrupt execution and trigger braking or replanning.

When the maneuver finishes, the runtime reassesses nominal proposals.
Nominal control resumes when a proposal is admitted; otherwise, recovery
planning is invoked again.
If no admissible trajectory is available, the runtime applies conservative
braking as a contingency response.
This fallback is not conditioned on passing the PDMS admission threshold
and does not carry the same admission assurance as a verified candidate.

%% file: sec/appendix/e_rule_based_verifier.tex
\section{Rule-based Verifier}
\label{app:rules}

The rule-based verifier evaluates nominal proposals and recovery trajectories using the scene context supplied by the runtime.
This section specifies the scoring rules and admission criterion, then evaluates threshold sensitivity in closed-loop driving.
Appendix~\ref{app:experimental-details} documents the experimental configuration and evaluation protocols.

\paragraph{Plan representation.}
The expert trajectory representation described here consists of 8 poses at 0.5\,s intervals.
The verifier densifies the plan to 0.1\,s and adjusts its arc-length progress using the measured ego speed and acceleration and braking bounds of 3.5 and 4.0\,m/s$^2$, respectively.
This adjustment limits the progress assigned to sample $k$ at time $k\cdot 0.1$\,s according to these longitudinal bounds.
The adjusted trajectory is used for safety checks, while ego progress is evaluated on the original proposal by default.

\subsection{Scoring Function and Rule Definitions}
\label{app:pdms}

\paragraph{Score.}
Our rule-based verifier adapts the PDM Score of NavSim~\citep{navsim} to live CARLA state.
For a candidate plan $\tau$, it computes
\begin{equation}
\mathrm{PDMS}(\tau)
=
\mathrm{NC}
\cdot \mathrm{DAC}
\cdot \mathrm{TTC}_{\mathrm{VRU}}
\cdot
\frac{5\,\mathrm{TTC}+5\,\mathrm{EP}+2\,\mathrm{LK}}{12},
\end{equation}
where NC and $\mathrm{TTC}_{\mathrm{VRU}}$ are binary gates, DAC is a tiered gate in $\{0,0.5,1\}$, and TTC, EP, and LK are soft terms in $[0,1]$.
TTC and LK take binary values in the implementation, whereas EP varies continuously.
The same function scores candidate batches during trajectory selection.

\paragraph{Gates.}
No collisions (NC) sweeps the ego bounding box along the adjusted plan against all detected agents.
Vehicles are forecast with the kinematic bicycle model of World on Rails~\citep{chen2021wor}, with the CARLA parameters used by TransFuser++~\citep{jaeger2023hidden}, under held controls.
Walkers classified as moving using a 0.3\,m/s speed threshold are forecast with a minimum speed of 1.5\,m/s along their heading, while stationary walkers are held at their measured pose.
Collisions within the configured veto horizons zero the NC gate: 1.5\,s for vehicles and static objects and 3.0\,s for pedestrians, each extended to the ego stopping time plus 0.5\,s when that is longer.
Drivable area compliance (DAC) queries the HD map at the checked plan endpoint and assigns a tier of 1.0 for any driving or parking lane, 0.5 on the shoulder, and 0.0 for off-road locations or sidewalks.
The VRU veto ($\mathrm{TTC}_{\mathrm{VRU}}$) zeros the score when the TTC term fails because of a pedestrian.
This additional gate prevents a pedestrian-related TTC failure from being compensated by the remaining soft terms, which could otherwise yield a score as high as $7/12 \approx 0.583$ and pass the admission threshold.

\paragraph{Soft terms.}
Time to collision (TTC) probes the plan at forward offsets of 0.2, 0.3, 0.5, 0.8, and 1.0\,s, with surrounding agents predicted at the corresponding future times.
An overlap within the applicable veto horizon at any of these offsets sets TTC to 0.
Ego progress (EP) is the arc-length progress of the original plan along the route, divided by the progress expected at 5\,m/s over the plan horizon and clipped to $[0,1]$.
Lane keeping (LK) is 0 when the checked plan endpoint is more than 1.5\,m from the nearest driving lane centerline, and 1 otherwise.

\paragraph{Deviations from NavSim.}
Comfort was replaced with lane keeping (LK) because executed accelerations are set by the controller rather than the plan.
Our drivable area compliance (DAC) calculation is tiered because CARLA exposes lane types rather than drivable area polygons.
Collision gates are limited to a veto horizon instead of the full plan, and vehicles are forecast with a bicycle model instead of constant velocity.
We add a VRU veto to place an emphasis on avoiding pedestrian collisions.

\subsection{Admission Criterion and Score Bounds}
\label{app:verifier-admission}

\paragraph{Admission and runtime integration.}
A plan passes the score-based admission test when $\mathrm{PDMS}\geq\theta$, with $\theta=0.43$ in the default configuration.
Otherwise, the plan is rejected.
When no nominal candidate is admitted, control passes to recovery planning when enabled, with controlled fallback used when no executable proposal is available.
The runtime control flow is specified in Section~\ref{app_sec:runtime-control-flow}, and recovery execution, interruption, and fallback are described in Section~\ref{app_sec:recovery-execution}.
Route-consistency checks impose additional conditions during candidate
selection and are not terms in the PDMS equation.
After a recovery trajectory is committed, maneuver-specific execution
guards monitor its continuation using updated observations; these guards
are distinct from the full admission check used during candidate selection.

\paragraph{Threshold choice.}
With all multiplicative gates equal to one, setting any two of TTC, EP, and LK to zero limits PDMS to at most $5/12 \approx 0.417$; the default threshold $\theta=0.43$ therefore rejects all such combinations.
A single zero-valued term does not necessarily cause rejection: the score can reach $7/12 \approx 0.583$ when TTC or EP is zero, and $10/12 \approx 0.833$ when LK is zero.
Thresholds above these boundaries make the corresponding soft terms mandatory for admission; reduced multiplicative gates can only lower the score.
These bounds explain the admission rule but do not establish an optimal closed-loop threshold.
Section~\ref{app_sec:threshold_sensitivity} evaluates the empirical trade-off between collisions and timeouts.

\paragraph{Scope of admission.}
Admission denotes satisfaction of the configured rules under the supplied
scene estimate and motion forecasts, within the configured checking horizons.
It does not require every soft term to be nonzero:
a vehicle-related TTC failure can be compensated by the remaining soft
terms when all multiplicative gates permit admission, whereas an NC
failure or a pedestrian-related TTC failure vetoes the candidate.
In the SparseDriveV2 configuration, DAC and LK evaluate only the checked
endpoint, so passing these checks does not establish map compliance at
intermediate trajectory positions.
These conditions do not establish continuous collision avoidance or
drivable-area compliance over the entire trajectory or its subsequent
closed-loop execution.

\subsection{Verifier Threshold Sensitivity}
\label{app_sec:threshold_sensitivity}

Table~\ref{tab:threshold-sweep} evaluates six thresholds on 34 Fail2Drive routes using the full \ourmethod runtime with SparseDriveV2.
DS and SR are highest at $\theta=0.43$ among the tested values.
Reducing the threshold to 0.01 raises collisions per route from 0.059 to 0.176, while increasing it to 0.60 or 0.84 reduces collisions but raises the number of timeouts from 2 to 8 or 12, respectively.
Stricter admission therefore does not monotonically improve task completion: lower collision rates can coincide with more stalled routes.
These results support the default threshold of 0.43 on this evaluated subset, without establishing its optimality across other scenarios.

\begin{table}[t]
    \centering
    \caption{\textbf{Verifier threshold sweep (17 scenario families).} One Base/Generalization
    route pair per Fail2Drive scenario family (34 routes): 17 pairs of routes are drawn uniformly at random within their
    families. Full \ourmethod runtime with SparseDriveV2, 180\,s simulation cap; routes reaching the cap are scored as scenario timeouts (Appendix~\ref{sec:fail2drive-setup}).
    HM is the harmonic mean of DS and SR. Timeouts counts routes that reach the simulation cap.
    Rejected is the fraction of control ticks at which the verifier rejects the selected proposal
    and triggers recovery.}
    \label{tab:threshold-sweep}
    \small
    \renewcommand{\arraystretch}{1.12}
    \setlength{\tabcolsep}{5pt}
    \begin{tabular}{@{}lcccccc@{}}
        \toprule
        Threshold & DS $\uparrow$ & SR (\%) $\uparrow$ & HM $\uparrow$ & Col. $\downarrow$ & Timeouts $\downarrow$ & Rejected (\%) \\
        \midrule
        0.01 (hard gates only)          & 75.0 & 70.6 & 72.7 & 0.176 & 4  & 1.8 \\
        0.40                            & 79.5 & 85.3 & 82.3 & 0.118 & 2  & 0.7 \\
        \textbf{0.43 (ours)}            & \textbf{81.1} & \textbf{88.2} & \textbf{84.5} & 0.059 & 2  & 1.1 \\
        0.45                            & 76.2 & 82.4 & 79.1 & 0.147 & 2  & 1.0 \\
        0.60 (TTC becomes a hard gate)  & 73.3 & 76.5 & 74.9 & 0.029 & 8  & 4.2 \\
        0.84 (any soft defect rejects)  & 64.5 & 64.7 & 64.6 & 0.000 & 12 & 8.6 \\
        \bottomrule
    \end{tabular}
\end{table}

%% file: sec/appendix/f_experimental_protocols.tex
\section{Experimental Protocols and Reproducibility}
\label{app:experimental-details}

\subsection{Models, Sensors, and Hardware}
\label{app:models-runtime-config}

\paragraph{Driving models.}
The runtime integrates pretrained SparseDriveV2, UniAD, and BridgeDrive through model-specific adapters.
The SparseDriveV2 configuration uses the Bench2Drive model defined by \texttt{projects/configs/sparsedrive\_stage2.py} and the checkpoint \texttt{sparsedrive\_small\_b2d\_stage2.pth}.
The UniAD adapter uses the Bench2Drive configuration \texttt{base\_e2e\_b2d.py} and checkpoint \texttt{uniad\_base\_b2d.pth}.
The BridgeDrive adapter loads the \texttt{bridgedrive\_k60} checkpoint directory through the LEAD implementation.
Model inference is separated from the CARLA agent through worker processes, allowing each model to use its own software environment.

\paragraph{Vision-language models.}
The local VLM configuration specifies \texttt{Qwen/Qwen3.5-9B}.
Separate API configurations specify \texttt{claude-opus-5} and \texttt{gpt-6-astra} for the Opus~5 and GPT-6 Astra variants, respectively.
These recovery configurations use the front-camera view.
Prompt construction, temporal image inputs, and recovery invocation are described in Section~\ref{app_sec:recovery-vlm-protocol}.

\paragraph{Sensor inputs and perception.}
SparseDriveV2 uses its fixed six-camera Bench2Drive rig, covering the front, front-left, front-right, rear, rear-left, and rear-right views.
In the SparseDriveV2 runtime configuration, the shared adapted BridgeDrive detector runs as a separate service with an additional front RGB camera, two LiDAR sensors, four radar sensors, IMU, and GNSS inputs.
The additional front camera has a resolution of $384\times384$ and a field of view of $60^\circ$.
The reduced perception rig supplies the central BridgeDrive camera view and fills the two unavailable side-camera inputs with zeros.
A separate configuration option enables all three BridgeDrive camera views.
When BridgeDrive is itself the driving model, its adapter registers the three native RGB cameras at yaw angles of $-54.5^\circ$, $0^\circ$, and $54.5^\circ$.
Each of these cameras has a resolution of $384\times384$ and a field of view of $60^\circ$.

The SparseDriveV2 sensor-based configuration obtains surrounding-object detections from BridgeDrive and enables radar- and LiDAR-based box refinement.
It disables the configured sources of simulator object annotations while retaining HD-map lane information for verification and route handling.
The perception confidence threshold is 0.35, inference is scheduled every four control ticks, and the maximum permitted detection age is 1.0\,s.
The scene-refinement procedures are described in Section~\ref{app:scene-context-refinement}.
The traditional monitors use the same adapted BridgeDrive detector; their intervention mechanisms are specified in Section~\ref{app:traditional-monitors}.

On Fail2Drive, the runtime uses the same adapted BridgeDrive detector when paired with SparseDriveV2, UniAD, or BridgeDrive as the driving expert.
Simulator ground-truth scene context is used only in the explicitly labeled oracle ablation.

\paragraph{Perception adaptation.}
The pedestrian and static-obstacle modules are trained on CARLA sensor observations with simulator-generated object annotations; the original BridgeDrive detector and driving expert remain frozen.
Both modules use dual-LiDAR inputs, and the static-obstacle module additionally uses frozen BridgeDrive BEV features.
Pedestrian predictions replace the base detector's pedestrian outputs, while static-obstacle proposals are merged with base detections using non-maximum suppression; vehicle predictions remain unchanged.
Postprocessing also applies local box-center refinement and pedestrian shape filtering.
We generate 1,831 and 2,520 annotated sensor frames in CARLA to train the pedestrian and static-obstacle modules, respectively.
These modules are used in the Fail2Drive sensor-based runtime evaluations, including the traditional-monitor comparison, but not on NavSafe-$\infty$.
The closed-loop evaluations do not separately isolate the contribution of detector adaptation.

\paragraph{Control and candidate evaluation.}
The SparseDriveV2 configuration uses a 20\,Hz control and replanning schedule in simulated time.
This does not imply a wall-clock throughput of 20\,Hz.
Its candidate scorer uses a default top-$30$ native-proposal branch and enables a separate region-sampling branch with up to three proposals per region and sampling seed 0.
The region-sampling seed is distinct from the simulator evaluation seed.
The SparseDriveV2 configuration selects candidates using PDMS ranking rather than enabling the optional native-ranking-after-PDMS-gating mode.
The BridgeDrive configuration instead sets the native-proposal limit to 60 and enables native ranking among proposals that pass the PDMS gate.
The UniAD configuration produces six poses at 0.5\,s intervals and performs a new model and orchestration pass every five control ticks.
Between these passes, its cached trajectory is reprojected and checked on each control tick.

\paragraph{Verifier configuration.}
Table~\ref{tab:verifier-runtime-config} records the verifier settings shared by the evaluated SparseDriveV2 sensor-based configurations, including the local-VLM, Opus~5, and GPT-6 Astra variants.
The definitions of the scoring terms are given in Appendix~\ref{app:rules}, and the threshold sensitivity experiment is reported in Section~\ref{app_sec:threshold_sensitivity}.

\begin{table}[t]
    \centering
    \caption{\textbf{Verifier configuration for the SparseDriveV2 sensor-based setting.}
    Values are shared by the evaluated local-VLM, Opus~5, and GPT-6 Astra configurations.}
    \label{tab:verifier-runtime-config}
    \small
    \renewcommand{\arraystretch}{1.12}
    \setlength{\tabcolsep}{5pt}
    \begin{tabular}{@{}ll@{}}
        \toprule
        Parameter & Value \\
        \midrule
        Admission threshold & 0.43 \\
        Verification sampling interval & 0.1\,s \\
        Vehicle/static-object NC horizon & 1.5\,s \\
        Pedestrian veto horizon & 3.0\,s \\
        Stop-conditioned horizon extension & Enabled \\
        Reaction-time margin & 0.5\,s \\
        TTC probe offsets & $\{0.2,0.3,0.5,0.8,1.0\}$\,s \\
        TTC debounce length & 1 check \\
        Pedestrian TTC veto & Enabled \\
        Vehicle forecast model & Kinematic bicycle \\
        Moving-pedestrian minimum forecast speed & 1.5\,m/s \\
        Additional stationary-vehicle lateral margin & 0.0\,m \\
        \midrule
        HD-map evaluation mode & Endpoint only \\
        DAC mode & Tiered \\
        Configured DAC horizon & 1.0\,s \\
        Configured DAC exit grace & Enabled \\
        LK deviation limit & 1.5\,m \\
        Configured LK window & 2.0\,s \\
        \midrule
        Re-timing acceleration bound & 3.5\,m/s$^2$ \\
        Re-timing braking bound & 4.0\,m/s$^2$ \\
        EP evaluated on re-timed trajectory & Disabled \\
        Additional normal-mode EP gate & Disabled \\
        \bottomrule
    \end{tabular}
\end{table}

The collision-checking horizons are extended when the current stopping time plus the 0.5\,s reaction margin exceeds their configured values.
Endpoint-only HD-map evaluation passes only the final checked position to the DAC and LK calculations.
Consequently, the configured DAC horizon, DAC exit-grace option, and LK window do not provide an additional along-trajectory check in this mode.
The UniAD configuration instead disables endpoint-only HD-map evaluation.
The BridgeDrive configuration sets an additional stationary-vehicle lateral margin of 1.0\,m, whereas the SparseDriveV2 configuration uses 0.0\,m.
All three configurations use the same numerical admission threshold.

\paragraph{Route-instruction settings.}
The evaluated configurations enable the route-instruction scorer.
Its endpoint-bearing classifier uses a straight-heading threshold of $10^\circ$, a turn-heading threshold of $15^\circ$, and a minimum forward displacement of 0.3\,m.
The scorer assigns weights of 1.0, 0.5, and 0.0 to matching, ambiguous, and contradictory directions, respectively.
These settings belong to candidate selection and are separate from the PDMS soft-term weights.

\paragraph{Simulation and Hardware.}
We use the Fail2Drive distribution of CARLA 0.9.15 on a server equipped with eight NVIDIA L40S GPUs.
Each concurrent route slot runs a dedicated CARLA server and evaluator, with GPU assignment and the number of simultaneous slots configured by the evaluation launcher.
Local driving and perception models run on the assigned GPUs, while API-based VLM variants use external inference services.
The execution conditions for the runtime comparison are detailed in Appendix~\ref{app:runtime}.

\subsection{Fail2Drive Setup and Metrics}
\label{sec:fail2drive-setup}

Fail2Drive~\citep{fail2drive} contains 100 paired routes covering 17 unseen long-tail
scenario types.  Each pair consists of an in-distribution base route and a shifted
route with matched start, goal, route geometry, and non-target factors.  We evaluate
all 200 routes for the main result and report the paired generalization gap.

Our primary metric is the official driving score (DS), defined by the benchmark as
route completion (RC, in percent) multiplied by the infraction penalty (IP). We also report route success rate (SR), where a route is successful only if the evaluator records no infraction other than minimum-speed and outside-route-lane events, so collisions, traffic-rule violations, scenario timeouts, blockages, route timeouts, and route deviations all count as failures.
We report collision events per route (Col.) and HM, the harmonic mean of DS and SR. In Tables~\ref{tab:main-results} and~\ref{tab:runtime-comparison}, HM is computed within each split, and Average is the arithmetic mean of the two splits for each metric. DS measures the safety--progress trade-off, RC
separates progress from penalties, and SR exposes catastrophic closed-loop failures
that can be obscured by averages.
Each route is evaluated for at most 180\,s of simulated time to bound evaluation cost.
Routes that reach this limit without completing were stationary or near-stationary, and we score them as scenario timeouts: they count as unsuccessful for SR, and the benchmark's scenario-timeout penalty ($\times 0.7$) is applied to their infraction penalty when computing DS.
For recovery analysis, we report the recovery success rate (Rec.-SR) over temporally grouped recovery events: consecutive retries form one event, which succeeds when a verifier-admitted action completes and nominal driving then remains healthy for five consecutive simulation seconds.

For SparseDriveV2, the controller and planner run at 20\,Hz; UniAD instead runs a full model pass every five control ticks and re-verifies its cached trajectory on each tick. BridgeDrive is evaluated every four control ticks with a score threshold of $0.35$ and a maximum detection staleness of $1$\,s. In the SparseDriveV2 sensor-based configuration, the end-to-end proposal branch retains the expert's top 30 candidates.
For this configuration, the PDMS admission threshold is $0.43$; the longitudinal re-timing bounds are $3.5\,\mathrm{m/s^2}$ for acceleration and $4.0\,\mathrm{m/s^2}$ for braking; the NC gate horizon is $1.5$\,s for vehicles and static objects and $3.0$\,s for pedestrians, each extended to the ego stopping time plus $0.5$\,s when that is longer; DAC and LK are evaluated at the checked plan endpoint; and TTC is probed at $0.2$, $0.3$, $0.5$, $0.8$, and $1.0$\,s.

\subsection{NavSafe-$\infty$ Setup and Metrics}
\label{app:navsafe-metrics}

\paragraph{Driving experts and perception source.}
On NavSafe-$\infty$, we evaluate runtime augmentation of LTF, DiffusionDrive, and SparseDriveV2.
For all three augmented experts, the runtime obtains scene observations from BridgeDrive's native perception module.
The additional pedestrian and static-obstacle modules are not used in these NavSafe-$\infty$ experiments.
BridgeDrive supplies perception to the runtime, while the corresponding driving expert supplies nominal driving proposals.

For completeness we restate the metric definitions of NavSafe-$\infty$~\citep{bao2026Navsafe} used in \Cref{tab:navsafe}. We refer the reader to
  the original paper for the event specification, the semi-reactive traffic
  protocol, and the validity checks that determine which rollouts are scored.

 Each event begins after a two-second log-replay warm-up and runs until success,
  task failure, or timeout. Every policy is evaluated on the same preregistered seeds
  per event specification, and we report the mean across seeds. For a
  specification $b$, let $\mathcal{V}_b$ denote its rollouts that satisfy the
  benchmark's validity checks, and let
  $\mathcal{B}^{+} = \{b : |\mathcal{V}_b| > 0\}$ be the specifications retaining at
  least one such rollout. Metrics average first over the rollouts in
  $\mathcal{V}_b$ and then over the specifications in $\mathcal{B}^{+}$, so that
  every specification contributes equally to the reported score.

  \paragraph{Driving Score (DS).} DS combines task-specific completion
  $\mathrm{EC}_{br} \in [0,1]$ with multiplicative penalties $p_q \in [0,1]$ for
  recorded infractions,
  \begin{equation}
  \label{eq:navsafe-ds}
  \mathrm{DS} = \frac{100}{|\mathcal{B}^{+}|}
  \sum_{b \in \mathcal{B}^{+}} \frac{1}{|\mathcal{V}_b|}
  \sum_{r \in \mathcal{V}_b} \mathrm{EC}_{br} \prod_{q \in \mathcal{I}_{br}} p_q .
  \end{equation}
  Where partial completion is not meaningful for a task, $\mathrm{EC}_{br}$ is
  binary. Because the penalties multiply, one severe infraction bounds the score
  irrespective of how much of the task was completed.

  \paragraph{Success Rate (SR).} $s_{br} = 1$ only when rollout $r$ satisfies its
  success predicate within the time budget and incurs no disallowed infraction.
  With $\mathrm{SR}_b = |\mathcal{V}_b|^{-1} \sum_{r \in \mathcal{V}_b} s_{br}$,
  the benchmark score is
  $\mathrm{SR} = 100 |\mathcal{B}^{+}|^{-1} \sum_{b \in \mathcal{B}^{+}} \mathrm{SR}_b$.
  SR admits no partial credit and therefore exposes categorical failures that DS
  can average away.

  \paragraph{Driving Efficiency (DE).} DE is the ratio of ego speed to
  nearby-traffic speed.

  \paragraph{Comfort.} Comfort is the fraction of segments satisfying the
  acceleration, jerk, yaw-rate, and yaw-acceleration bounds. DE and Comfort use
  the same hierarchical averaging as DS and SR and are undefined when their
  denominator is empty.

  \paragraph{Ability scores (AS).} Specification-level success rates are grouped                                                              
  by the four RoadSafe365 categories: Traffic Crashes (TC), Vulnerable Road User
  Crashes (VRUC), Traffic Violations (TV), and Traffic Incidents
  (TI)~\citep{RoadSafe365}. Writing
  $\mathcal{B}^{+}_k = \{b \in \mathcal{B}^{+} : p(\ell_b) = k\}$ for the
  specifications whose taxonomy leaf $\ell_b$ has parent category $k$,
  \begin{equation}
  \label{eq:navsafe-as}
  \mathrm{AS}_k = \frac{100}{|\mathcal{B}^{+}_k|}
  \sum_{b \in \mathcal{B}^{+}_k} \mathrm{SR}_b ,
  \qquad k \in \{\mathrm{TC}, \mathrm{VRUC}, \mathrm{TV}, \mathrm{TI}\} .
  \end{equation}
  AS therefore inherits SR's binary criterion: progress, efficiency, comfort, and
  partial penalties cannot raise it. \Cref{tab:navsafe} reports the four category-specific ability scores.

\subsection{Alternative Runtime Implementation Details}
\label{app:traditional-monitors}

\paragraph{Common setup.}
All conditions execute the same frozen SparseDriveV2 checkpoint at 20\,Hz and use the same adapted BridgeDrive detector.
The detector provides surrounding-agent detections every four control ticks with a score threshold of $0.35$ and a maximum detection age of $1$\,s.
For all six alternative runtimes, corrective proposals replace the nominal trajectory only when they pass verification and achieve a higher PDMS score than the nominal proposal.
The 2D MPC and AESOP configurations additionally use their own constrained planning and execution procedures.

\paragraph{Evaluation routes.}
The uncertainty monitor requires threshold calibration. We reserve 40 routes for calibration and exclude them from the comparison for all methods. We evaluate all remaining 160 routes, comprising 80 In-Distribution/Generalization route pairs covering all 17 Fail2Drive scenario families. All compared methods use the same routes and seed~0, with one rollout per route. SparseDriveV2 and \ourmethod results are restricted to the same evaluation routes. Metrics are averaged separately over the 80 routes in each split, weighting each route equally.

\paragraph{RSS.}
The RSS wrapper evaluates the current longitudinal and lateral separation from each detected agent using a $1$\,s response time, maximum ego acceleration of $2.0\,\mathrm{m/s^2}$, minimum ego braking of $4.0\,\mathrm{m/s^2}$, assumed leading-agent braking of $8.0\,\mathrm{m/s^2}$, and $0.2$\,m minimum lateral clearance. An unsafe state triggers a bounded-braking proposal, with a stop proposed below $0.3\,\mathrm{m/s}$; the braking request is released after three consecutive safe updates.

\paragraph{MPC.}
The MPC wrapper assumes constant-velocity motion for detected agents and optimizes a longitudinal acceleration sequence over the native waypoint horizon at $0.5$\,s resolution. SLSQP minimizes path-progress, speed, control-effort, and acceleration-smoothness costs subject to acceleration in $[-4.0,2.4]\,\mathrm{m/s^2}$, speed below $12\,\mathrm{m/s}$, and sensed-agent clearance constraints. It preserves the native path geometry and proposes
$4.0\,\mathrm{m/s^2}$ braking when optimization is infeasible.

\paragraph{Uncertainty monitor.}
The uncertainty monitor combines normalized entropy and the relative top-two score margin from the expert's top 32 trajectory candidates.
The two components receive equal weight.
A calibrated threshold of approximately $0.923$ triggers intervention after three consecutive high-risk updates, with release after five low-risk updates.
The monitor initially proposes deceleration at $2.5\,\mathrm{m/s^2}$ along the nominal path, followed by a stop if the request persists for ten updates or speed falls below $0.3\,\mathrm{m/s}$.
This score is a failure-detection heuristic rather than a calibrated failure probability.

\paragraph{2D MPC.}
The 2D MPC jointly optimizes acceleration and front-wheel steering using a kinematic bicycle model.
The expert trajectory serves as a soft tracking reference, allowing the optimized path to depart laterally from it.
Constraints include vehicle dynamics, constant-velocity predictions of detected objects, and an HD-map corridor containing permitted same-direction driving lanes.
SLSQP optimizes controls at $0.5$\,s intervals, with intermediate collision checks and finer trajectory validation before execution.
Once admitted by verification and PDMS arbitration, the resulting trajectory is tracked by the SparseDriveV2 PID controller.
Missing planning context or failure to find a feasible trajectory triggers braking.

\paragraph{AESOP (adapted).}
Our CARLA adaptation follows AESOP's anomaly-triggered reasoning and branch-MPC structure~\citep{sinha2024realtimeanomalydetectionreactive}.
A MiniLM encoder compares a textual description of the sensed scene with a nominal embedding cache.
An anomaly triggers an asynchronous Qwen3.5-VL-9B request to select nominal driving or a terminal stop region.
The adaptation uses a bicycle-model branch MPC with six $0.5$\,s planning steps and a $1$\,s simulation-time reasoning-latency bound.
Nominal and fallback branches share their first control input, while fallback branches share a control prefix covering the reasoning delay.
Once the proposal passes verification and PDMS arbitration, its selected control is executed through the Ackermann controller.
Infeasible optimization, invalid reasoning output, or an exceeded latency bound triggers braking.
The adaptation uses finite-horizon stop regions and does not establish invariant-set safety guarantees.

\paragraph{AESOP + primitives.}
This variant retains the same anomaly detector, reasoning backend, branch MPC, and controller.
The language model instead selects nominal driving or an available primitive from our library.
The selected primitive is instantiated using its default, potentially scene-dependent parameters and supplied as a tracking reference to branch MPC.
The reference may therefore be modified or rejected by the optimizer before execution.
The language model does not generate numerical parameters or action sequences.
This variant does not enable \ourmethod's parameter search, verifier-guided primitive revision, or primitive execution controller.

\paragraph{Metrics and validation.}

The driving-score gap is $\mathrm{Gap}=\mathrm{DS}_{\mathrm{ID}}-\mathrm{DS}_{\mathrm{Gen}}$, computed from unrounded means and expressed in score points. Positive values indicate lower driving scores on generalization routes. The gap must be interpreted alongside both scores, since a smaller gap can also result from worse in-distribution performance.

\paragraph{Validation.}
Accepted runs must have complete evaluator results, matching configuration and provenance records, and telemetry confirming that the native SparseDriveV2 argmax is preserved before intervention. Valid driving failures, including blockage and route deviation, remain in the evaluation. Infrastructure-interrupted attempts are retained for review and are not counted as completed evaluations.

\subsection{Recovery Metrics and Statistical Aggregation}
Recovery events begin when a recovery action executes and group retries until nominal driving remains healthy for five consecutive simulation seconds. Success requires an action admitted with a verifier score of at least 0.43 to complete, followed by this interval of verifier-approved nominal driving without renewed recovery. Routes without recovery and events without a verifier-admitted action are excluded. Insufficient evidence or termination before confirmation yields an unknown outcome. We compute $\mathrm{Rec.\text{-}SR} = 100 \times N_{\mathrm{success}}/(N_{\mathrm{success}} + N_{\mathrm{failure}})$, excluding unknown outcomes. This measures sustained return to nominal planning; executor completion includes execution-budget expiry and does not independently certify geometric maneuver completion or collision-free recovery.

%% file: iclr2026_conference.bib
@article{agarwal2026cosmos,
  author  = {Agarwal, Niket and Ali, Arslan and Allen, Jon and Antolini, Martin and Aubame, Adeline and Azzolini, Alisson and Bai, Junjie and Bala, Maciej and Balaji, Yogesh and Bapst, Josh and others},
  journal = {arXiv preprint arXiv:2606.02800},
  title   = {{Cosmos 3}: Omnimodal World Models for Physical {AI}},
  year    = {2026}
}

@article{ali2025world,
  author =        {Ali, Arslan and Bai, Junjie and Bala, Maciej and
                   Balaji, Yogesh and Blakeman, Aaron and Cai, Tiffany and
                   Cao, Jiaxin and Cao, Tianshi and Cha, Elizabeth and
                   Chao, Yu-Wei and others},
  journal =       {arXiv preprint arXiv:2511.00062},
  title = {World Simulation with Video Foundation Models for Physical {AI}},
  year =          {2025},
}

@article{li2024bevformer,
  author  = {Li, Zhiqi and Wang, Wenhai and Li, Hongyang and Xie, Enze and Sima, Chonghao and Lu, Tong and Yu, Qiao and Dai, Jifeng},
  journal = {IEEE Transactions on Pattern Analysis and Machine Intelligence},
  title   = {{BEVFormer}: Learning Bird's-Eye-View Representation from {LiDAR}-Camera via Spatiotemporal Transformers},
  volume  = {47},
  number  = {3},
  pages   = {2020--2036},
  year    = {2025}
}

@article{liu2025hybrid,
  author =        {Liu, Haochen and Huang, Zhiyu and Huang, Wenhui and
                   Yang, Haohan and Mo, Xiaoyu and Lv, Chen},
  journal =       {IEEE Transactions on Pattern Analysis and Machine
                   Intelligence},
  number =        {4},
  pages =         {2597--2614},
  publisher =     {IEEE},
  title =         {Hybrid-prediction integrated planning for autonomous
                   driving},
  volume =        {47},
  year =          {2025},
}

@inproceedings{uniad,
  author =        {Hu, Yihan and Yang, Jiazhi and Chen, Li and Li, Keyu and
                   Sima, Chonghao and Zhu, Xizhou and Chai, Siqi and
                   Du, Senyao and Lin, Tianwei and Wang, Wenhai and
                   Lu, Lewei and Jia, Xiaosong and Liu, Qiang and
                   Dai, Jifeng and Qiao, Yu and Li, Hongyang},
  booktitle =     {Proceedings of the IEEE/CVF Conference on Computer
                   Vision and Pattern Recognition (CVPR)},
  title =         {Planning-oriented Autonomous Driving},
  year =          {2023},
  url =           {https://arxiv.org/abs/2212.10156},
}

@inproceedings{zhou2025v2xpnp,
  author =        {Zhou, Zewei and Xiang, Hao and Zheng, Zhaoliang and
                   Zhao, Seth Z and Lei, Mingyue and Zhang, Yun and
                   Cai, Tianhui and Liu, Xinyi and Liu, Johnson and
                   Bajji, Maheswari and others},
  booktitle =     {2025 IEEE/CVF International Conference on Computer
                   Vision (ICCV)},
  organization =  {IEEE},
  pages =         {25399--25409},
  title = {{V2XPnP}: Vehicle-to-Everything Spatio-Temporal Fusion for Multi-Agent Perception and Prediction},
  year =          {2025},
}

@inproceedings{zhou2025turbotrain,
  author =        {Zhou, Zewei and Zhao, Seth Z and Cai, Tianhui and
                   Huang, Zhiyu and Zhou, Bolei and Ma, Jiaqi},
  booktitle =     {2025 IEEE/CVF International Conference on Computer
                   Vision (ICCV)},
  organization =  {IEEE},
  pages =         {4391--4402},
  title = {{TurboTrain}: Towards Efficient and Balanced Multi-Task Learning for Multi-Agent Perception and Prediction},
  year =          {2025},
}

@inproceedings{egostatus,
  author =        {Li, Zhiqi and Yu, Zhiding and Lan, Shiyi and
                   Li, Jiahan and Kautz, Jan and Lu, Tong and
                   Alvarez, Jose M.},
  booktitle =     {Proceedings of the IEEE/CVF Conference on Computer
                   Vision and Pattern Recognition (CVPR)},
  title =         {Is Ego Status All You Need for Open-Loop End-to-End
                   Autonomous Driving?},
  year =          {2024},
  url =           {https://arxiv.org/abs/2312.03031},
}

@inproceedings{xu2026wod,
  author    = {Xu, Runsheng and Lin, Hubert and Jeon, Wonseok and Feng, Hao and Zou, Yuliang and Sun, Liting and Gorman, John and Tolstaya, Kate and Tang, Sarah and White, Brandyn and Sapp, Ben and Tan, Mingxing and Hwang, Jyh-Jing and Anguelov, Dragomir},
  booktitle = {Proceedings of the IEEE/CVF Conference on Computer Vision and Pattern Recognition (CVPR)},
  title     = {{WOD-E2E}: {Waymo} Open Dataset for End-to-End Driving in Challenging Long-tail Scenarios},
  year      = {2026}
}

@inproceedings{bench2drive,
  author =        {Jia, Xiaosong and Yang, Zhenjie and Li, Qifeng and
                   Zhang, Zhiyuan and Yan, Junchi},
  booktitle =     {Advances in Neural Information Processing Systems
                   (NeurIPS) Datasets and Benchmarks Track},
  title =         {{Bench2Drive}: Towards Multi-Ability Benchmarking of
                   Closed-Loop End-To-End Autonomous Driving},
  year =          {2024},
  url =           {https://arxiv.org/abs/2406.03877},
}

@article{zhao2026bridgesim,
  author =        {Zhao, Seth Z. and Wang, Luobin and Ruan, Hongwei and
                   Bao, Yuxin and Chen, Yilan and Leng, Ziyang and
                   Ravichandran, Abhijit and He, Honglin and Zhou, Zewei and
                   Han, Xu and Peri, Abhishek and Huang, Zhiyu and
                   Desai, Pranav and Christensen, Henrik and Ma, Jiaqi and
                   Zhou, Bolei},
  journal =       {arXiv preprint arXiv:2604.10856},
  title =         {{BridgeSim}: Unveiling the {OL--CL} Gap in End-to-End
                   Autonomous Driving},
  year =          {2026},
  url =           {https://arxiv.org/abs/2604.10856},
}

@article{fail2drive,
  author =        {Gerstenecker, Simon and Geiger, Andreas and
                   Renz, Katrin},
  journal =       {arXiv preprint arXiv:2604.08535},
  title =         {{Fail2Drive}: Benchmarking Closed-Loop Driving
                   Generalization},
  year =          {2026},
  url =           {https://arxiv.org/abs/2604.08535},
}

@inproceedings{Kim_2026_CVPR,
  author =        {Kim, Jungho and Oh, Jiyong and Yu, Seunghoon and
                   Shin, Hongjae and Kwak, Donghyuk and Choi, Jun Won},
  booktitle =     {Proceedings of the IEEE/CVF Conference on Computer
                   Vision and Pattern Recognition (CVPR)},
  pages =         {24854--24864},
  title =         {SafeDrive: Fine-Grained Safety Reasoning for
                   End-to-End Driving in a Sparse World},
  year =          {2026},
}

@article{karkus2025beyond,
  author  = {Karkus, Peter and Igl, Maximilian and Chen, Yuxiao and Chitta, Kashyap and Packer, Jef and Douillard, Bertrand and Tian, Ran and Naumann, Alexander and Garcia-Cobo, Guillermo and Tan, Shuhan and De{\u{g}}irmenci, Alperen and Popov, Alexander and Smolyanskiy, Nikolai and Muller, Urs and Ivanovic, Boris and Pavone, Marco},
  journal = {TechRxiv preprint},
  title   = {Beyond Behavior Cloning in Autonomous Driving: A Survey of Closed-Loop Training Techniques},
  year    = {2025},
  doi     = {10.36227/techrxiv.176591565.57674132/v1},
  url     = {https://www.techrxiv.org/doi/full/10.36227/techrxiv.176591565.57674132/v1}
}

@misc{anthropic2025claudecode,
  author = {Anthropic},
  title  = {{Claude} Code: Best Practices for Agentic Coding},
  year   = {2025},
  url    = {https://www.anthropic.com/engineering/claude-code-best-practices},
  note   = {Accessed April 2026}
}

@article{liu2025robopilot,
  author =        {Liu, Xinyi and Sani, Mohammadreza Fani and
                   Zhou, Zewei and Wirbel, Julius and Zarrin, Bahram and
                   Galeazzi, Roberto},
  journal =       {arXiv preprint arXiv:2510.00154},
  title =         {RoboPilot: Generalizable Dynamic Robotic Manipulation
                   with Dual-thinking Modes},
  year =          {2025},
}

@article{zhang2026qwen,
  author =        {Zhang, Jiazhao and Zhou, Gengze and Yin, Hale and
                   Huang, Yiyang and Lei, Zixing and Peng, Qihang and
                   Yuan, Haoqi and Zhang, Jie and Guo, Xudong and
                   Chen, Xiaoyue and others},
  journal =       {arXiv preprint arXiv:2606.18112},
  title = {{Qwen-RobotNav} Technical Report: A Scalable Navigation Model Designed for an Agentic Navigation System},
  year =          {2026},
}

@article{bao2026Navsafe,
  author =        {Bao, Yuxin and Ruan, Hongwei and Wang, Luobin and
                   Zhao, Seth Z. and Leng, Ziyang and Zhang, Zihan and
                   Zeng, Yu and McAllister, Rowan and
                   Christensen, Henrik and Zhou, Bolei},
  journal =       {arXiv preprint arXiv:2609.26618},
  title =         {NavSafe-$\infty$: Benchmarking Closed-Loop Driving
                   Safety in Photorealistic Environments},
  year =          {2026},
  url =           {https://arxiv.org/abs/2609.26618},
}

@inproceedings{vad,
  author =        {Jiang, Bo and Chen, Shaoyu and Xu, Qing and
                   Liao, Bencheng and Chen, Jiajie and Zhou, Helong and
                   Zhang, Qian and Liu, Wenyu and Huang, Chang and
                   Wang, Xinggang},
  booktitle =     {Proceedings of the IEEE/CVF International Conference
                   on Computer Vision (ICCV)},
  title =         {{VAD}: Vectorized Scene Representation for Efficient
                   Autonomous Driving},
  year =          {2023},
  url =           {https://arxiv.org/abs/2303.12077},
}

@article{hydramdp,
  author =        {Li, Zhenxin and Li, Kailin and Wang, Shihao and
                   Lan, Shiyi and Yu, Zhiding and Ji, Yishen and
                   Li, Zhiqi and Zhu, Ziyue and Kautz, Jan and
                   Wu, Zuxuan and Jiang, Yu-Gang and Alvarez, Jose M.},
  journal =       {arXiv preprint arXiv:2406.06978},
  title =         {{Hydra-MDP}: End-to-end Multimodal Planning with
                   Multi-target Hydra-Distillation},
  year =          {2024},
  url =           {https://arxiv.org/abs/2406.06978},
}

@inproceedings{diffusiondrive,
  author =        {Liao, Bencheng and Chen, Shaoyu and Yin, Haoran and
                   Jiang, Bo and Wang, Cheng and Yan, Sixu and
                   Zhang, Xinbang and Li, Xiangyu and Zhang, Ying and
                   Zhang, Qian and Wang, Xinggang},
  booktitle =     {Proceedings of the IEEE/CVF Conference on Computer
                   Vision and Pattern Recognition (CVPR)},
  title =         {{DiffusionDrive}: Truncated Diffusion Model for
                   End-to-End Autonomous Driving},
  year =          {2025},
  url =           {https://arxiv.org/abs/2411.15139},
}

@article{sun2026sparsedrivev2,
  author =        {Sun, Wenchao and Lin, Xuewu and Chen, Keyu and
                   Pei, Zixiang and Li, Xiang and Shi, Yining and
                   Zheng, Sifa},
  journal =       {arXiv preprint arXiv:2603.29163},
  title = {{SparseDriveV2}: Scoring is All You Need for End-to-End Autonomous Driving},
  year =          {2026},
}

@inproceedings{zhou2026autovla,
  author    = {Zhou, Zewei and Cai, Tianhui and Zhao, Seth Z. and
               Zhang, Yun and Huang, Zhiyu and Zhou, Bolei and
               Ma, Jiaqi},
  booktitle = {Advances in Neural Information Processing Systems (NeurIPS)},
  title     = {{AutoVLA}: A Vision-Language-Action Model for End-to-End Autonomous Driving with Adaptive Reasoning and Reinforcement Fine-Tuning},
  volume    = {38},
  pages     = {27920--27956},
  year      = {2025},
}

@article{zhou2026spanvla,
  author =        {Zhou, Zewei and Yang, Ruining and Guo, Yiluan and
                   Chen, Sherry X and Feng, Tao and Pistunova, Kateryna and
                   Shen, Yishan and Su, Lili and Ma, Jiaqi},
  journal =       {arXiv preprint arXiv:2604.19710},
    title = {{SpanVLA}: Efficient Action Bridging and Learning from Negative-Recovery Samples for Vision-Language-Action Model},
  year =          {2026},
}

@inproceedings{li2025drivevla,
  author    = {Li, Yingyan and Shang, Shuyao and Liu, Weisong and Zhan, Bing and Wang, Haochen and Wang, Yuqi and Chen, Yuntao and Wang, Xiaoman and An, Yasong and Tang, Chufeng and Hou, Lu and Fan, Lue and Zhang, Zhaoxiang},
  title     = {{DriveVLA-W0}: World Models Amplify Data Scaling Law in Autonomous Driving},
  booktitle = {International Conference on Learning Representations (ICLR)},
  year      = {2026},
  url       = {https://arxiv.org/abs/2510.12796}
}

@article{worldengine,
  author =        {Li, Tianyu and Chen, Li and Wang, Caojun and
                   Liu, Haochen and Chitta, Kashyap and Yang, Zhenjie and
                   Lu, Yuhang and Ye, Naisheng and Qiu, Yihang and
                   Wang, Yufei and Zou, Luoxi and Peng, Jiaxin and
                   Pan, Jin and Su, Zhaoyu and Bursuc, Andrei and
                   Li, Shengbo Eben and Geiger, Andreas and Su, Peng and
                   Li, Hongyang},
  journal =       {arXiv preprint arXiv:2606.19836},
  title =         {{World Engine}: Towards the Era of Post-Training for
                   Autonomous Driving},
  year =          {2026},
  url =           {https://arxiv.org/abs/2606.19836},
}

@article{shalevshwartz2017rss,
  author =        {Shalev-Shwartz, Shai and Shammah, Shaked and
                   Shashua, Amnon},
  journal =       {arXiv preprint arXiv:1708.06374},
  title =         {On a Formal Model of Safe and Scalable Self-driving Cars},
  year =          {2017}
}

@article{mpcshield,
  author =        {Wabersich, Kim P. and Zeilinger, Melanie N.},
  journal =       {Automatica},
  title =         {A Predictive Safety Filter for Learning-Based Control
                   of Constrained Nonlinear Dynamical Systems},
  volume =        {129},
  year =          {2021},
  url =           {https://arxiv.org/abs/1812.05506}
}

@inproceedings{sinha2023fallback,
  author =        {Sinha, Rohan and Schmerling, Edward and
                   Pavone, Marco},
  booktitle =     {Proceedings of the IEEE Conference on Decision and
                   Control (CDC)},
  title =         {Closing the Loop on Runtime Monitors with
                   Fallback-Safe {MPC}},
  year =          {2023},
  doi =           {10.1109/CDC49753.2023.10383965},
  url =           {https://arxiv.org/abs/2309.08603},
}

@inproceedings{contreras2025soda,
  author =        {Contreras, Jose Leopoldo and Shorinwa, Ola and
                   Schwager, Mac},
  booktitle =     {Proceedings of the 7th Annual Learning for Dynamics
                   and Control Conference},
  pages =         {194--207},
  series =        {Proceedings of Machine Learning Research},
  title =         {Safe, Out-of-Distribution-Adaptive {MPC} with
                   Conformalized Neural Network Ensembles},
  volume =        {283},
  year =          {2025},
  url =           {https://proceedings.mlr.press/v283/contreras25a.html},
}

@inproceedings{yao2023react,
  author =        {Yao, Shunyu and Zhao, Jeffrey and Yu, Dian and
                   Du, Nan and Shafran, Izhak and Narasimhan, Karthik and
                   Cao, Yuan},
  booktitle =     {International Conference on Learning Representations
                   (ICLR)},
  title =         {{ReAct}: Synergizing Reasoning and Acting in Language
                   Models},
  year =          {2023},
  url =           {https://openreview.net/forum?id=WE_vluYUL-X},
}

@inproceedings{schick2023toolformer,
  author    = {Schick, Timo and Dwivedi-Yu, Jane and Dess{\`i}, Roberto and Raileanu, Roberta and Lomeli, Maria and Hambro, Eric and Zettlemoyer, Luke and Cancedda, Nicola and Scialom, Thomas},
  booktitle = {Advances in Neural Information Processing Systems (NeurIPS)},
  title     = {{Toolformer}: Language Models Can Teach Themselves to Use Tools},
  volume    = {36},
  year      = {2023},
  url       = {https://proceedings.neurips.cc/paper/2023/hash/d842425e4bf79ba039352da0f658a906-Abstract-Conference.html}
}

@inproceedings{wu2024autogen,
  author =        {Wu, Qingyun and Bansal, Gagan and Zhang, Jieyu and
                   Wu, Yiran and Li, Beibin and Zhu, Erkang and
                   Jiang, Li and Zhang, Xiaoyun and Zhang, Shaokun and
                   Liu, Jiale and Awadallah, Ahmed Hassan and
                   White, Ryen W. and Burger, Doug and Wang, Chi},
  booktitle =     {Conference on Language Modeling (COLM)},
  title =         {{AutoGen}: Enabling Next-Gen {LLM} Applications via
                   Multi-Agent Conversations},
  year =          {2024},
  url =           {https://openreview.net/forum?id=BAakY1hNKS},
}

@inproceedings{yang2024sweagent,
  author =        {Yang, John and Jimenez, Carlos E. and
                   Wettig, Alexander and Lieret, Kilian and Yao, Shunyu and
                   Narasimhan, Karthik and Press, Ofir},
  booktitle =     {Advances in Neural Information Processing Systems
                   (NeurIPS)},
  title =         {{SWE-agent}: Agent-Computer Interfaces Enable
                   Automated Software Engineering},
  volume =        {37},
  year =          {2024},
}

@article{wang2024voyager,
  author =        {Wang, Guanzhi and Xie, Yuqi and Jiang, Yunfan and
                   Mandlekar, Ajay and Xiao, Chaowei and Zhu, Yuke and
                   Fan, Linxi and Anandkumar, Anima},
  journal =       {Transactions on Machine Learning Research},
  title =         {{Voyager}: An Open-Ended Embodied Agent with Large
                   Language Models},
  year =          {2024},
  url =           {https://openreview.net/forum?id=ehfRiF0R3a},
}

@article{enpire,
  author =        {Xiao, Wenli and Xie, Jia and Zhang, Tonghe and
                   Lin, Haotian and Fu, Letian and Xue, Haoru and
                   Lu, Jalen and Yang, Yi and Dai, Cunxi and Wang, Zi and
                   Wu, Jimmy and Wang, Guanzhi and Sastry, S. Shankar and
                   Goldberg, Ken and Fan, Linxi and Zhu, Yuke and
                   Shi, Guanya},
  journal =       {arXiv preprint arXiv:2606.19980},
  title =         {{ENPIRE}: Agentic Robot Policy Self-Improvement in
                   the Real World},
  year =          {2026},
  url =           {https://arxiv.org/abs/2606.19980},
}

@article{aspire,
  author =        {Lu, Runyu and Wu, Yubo and Kou, Ethan and Fu, Letian and
                   Xiao, Wenli and Mandlekar, Ajay and Xu, Yinzhen and
                   Shi, Guanya and Goldberg, Ken and Chen, Ang and
                   Chowdhury, Mosharaf and Zhu, Yuke and Fan, Linxi and
                   Wang, Guanzhi},
  journal =       {arXiv preprint arXiv:2607.00272},
  title =         {{ASPIRE}: Agentic /Skills Discovery for Robotics},
  year =          {2026},
  url =           {https://arxiv.org/abs/2607.00272},
}

@inproceedings{mao2024agentdriver,
  author =        {Mao, Jiageng and Ye, Junjie and Qian, Yuxi and
                   Pavone, Marco and Wang, Yue},
  booktitle =     {Conference on Language Modeling (COLM)},
  title =         {A Language Agent for Autonomous Driving},
  year =          {2024},
  url =           {https://arxiv.org/abs/2311.10813},
}

@article{chen2026avo,
  author =        {Chen, Terry and Ye, Zhifan and Xu, Bing and Ye, Zihao and
                   Liu, Timmy and Hassani, Ali and Chen, Tianqi and
                   Kerr, Andrew and Wu, Haicheng and Xu, Yang and
                   Chen, Yu-Jung and Chen, Hanfeng and Kane, Aditya and
                   Krashinsky, Ronny and Liu, Ming-Yu and Grover, Vinod and
                   Ceze, Luis and Bringmann, Roger and Tran, John and
                   Liu, Wei and Xie, Fung and Lightstone, Michael and
                   Shi, Humphrey},
  journal =       {arXiv preprint arXiv:2603.24517},
  title =         {{AVO}: Agentic Variation Operators for Autonomous
                   Evolutionary Search},
  year =          {2026},
  url =           {https://arxiv.org/abs/2603.24517},
}

@inproceedings{pdmclosed,
  author =        {Dauner, Daniel and Hallgarten, Marcel and
                   Geiger, Andreas and Chitta, Kashyap},
  booktitle =     {Proceedings of the 7th Conference on Robot Learning
                   (CoRL)},
  pages =         {1268--1281},
  publisher =     {PMLR},
  series =        {Proceedings of Machine Learning Research},
  title =         {Parting with Misconceptions about Learning-based
                   Vehicle Motion Planning},
  volume =        {229},
  year =          {2023},
  url =           {https://arxiv.org/abs/2306.07962},
}

@inproceedings{liu2025bridgedrive,
  author =        {Liu, Shu and Chen, Wenlin and Li, Weihao and
                   Wang, Zheng and Yang, Lijin and Huang, Jianing and
                   Zhang, Yipin and Huang, Zhongzhan and Cheng, Ze and
                   Yang, Hao},
  booktitle =     {International Conference on Learning Representations
                   (ICLR)},
  title =         {BridgeDrive: Diffusion Bridge Policy for Closed-Loop
                   Trajectory Planning in Autonomous Driving},
  year =          {2026},
  url =           {https://arxiv.org/abs/2509.23589},
}

@inproceedings{wu2022trajectory,
    author = {Wu, Penghao and Jia, Xiaosong and Chen, Li and
                   Yan, Junchi and Li, Hongyang and Qiao, Yu},
    title = {Trajectory-guided control prediction for end-to-end
                   autonomous driving: A simple yet strong baseline},
    booktitle = {Advances in Neural Information Processing Systems},
    year = {2022},
}

@inproceedings{fu2025orion,
  author =        {Fu, Haoyu and Zhang, Diankun and Zhao, Zongchuang and
                   Cui, Jianfeng and Liang, Dingkang and Zhang, Chong and
                   Zhang, Dingyuan and Xie, Hongwei and Wang, Bing and
                   Bai, Xiang},
  booktitle =     {Proceedings of the IEEE/CVF International Conference
                   on Computer Vision (ICCV)},
  pages =         {24823--24834},
  title = {{ORION}: A Holistic End-to-End Autonomous Driving Framework by Vision-Language Instructed Action Generation},
  year =          {2025},
  url =           {https://arxiv.org/abs/2503.19755},
}

@inproceedings{tang2025hipad,
  author =        {Tang, Yingqi and Xu, Zhuoran and Meng, Zhaotie and
                   Cheng, Erkang},
  booktitle =     {Proceedings of the IEEE/CVF International Conference
                   on Computer Vision (ICCV)},
  title = {{HiP-AD}: Hierarchical and Multi-Granularity Planning with Deformable Attention for Autonomous Driving in a Single Decoder},
  year =          {2025},
  url =           {https://arxiv.org/abs/2503.08612},
}

@inproceedings{renz2025simlingo,
  author =        {Renz, Katrin and Chen, Long and Arani, Elahe and
                   Sinavski, Oleg},
  booktitle =     {Proceedings of the IEEE/CVF Conference on Computer
                   Vision and Pattern Recognition (CVPR)},
  pages =         {11993--12003},
  title =         {SimLingo: Vision-Only Closed-Loop Autonomous Driving
                   with Language-Action Alignment},
  year =          {2025},
  url =           {https://arxiv.org/abs/2503.09594},
}

@article{wu2026aligndrive,
  author =        {Wu, Yanhao and Zhang, Haoyang and He, Fei and Wu, Rui and
                   Shan, Yanhu and Qiu, Congpei and Gao, Liang and
                   Ke, Wei and Zhang, Tong},
  journal =       {arXiv preprint arXiv:2601.01762},
  title =         {AlignDrive: Aligned Lateral-Longitudinal Planning for
                   End-to-End Autonomous Driving},
  year =          {2026},
  url =           {https://arxiv.org/abs/2601.01762},
}

@inproceedings{holtz2026bevad,
  author =        {Holtz, David and Hanselmann, Niklas and Doll, Simon and
                   Cordts, Marius and Schiele, Bernt},
  booktitle =     {Proceedings of the IEEE/CVF Conference on Computer
                   Vision and Pattern Recognition (CVPR) Findings},
  title =         {What Matters for Scalable and Robust Learning in
                   End-to-End Driving Planners?},
  year =          {2026},
  url =           {https://arxiv.org/abs/2603.15185},
}

@inproceedings{jaeger2023hidden,
  author =        {Jaeger, Bernhard and Chitta, Kashyap and
                   Geiger, Andreas},
  booktitle =     {Proceedings of the IEEE/CVF International Conference
                   on Computer Vision (ICCV)},
  title =         {Hidden Biases of End-to-End Driving Models},
  year =          {2023},
  url =           {https://arxiv.org/abs/2306.07957},
}

@inproceedings{lead2025,
  author =        {Nguyen, Long and Fauth, Micha and Jaeger, Bernhard and
                   Dauner, Daniel and Igl, Maximilian and
                   Geiger, Andreas and Chitta, Kashyap},
  booktitle =     {Proceedings of the IEEE/CVF Conference on Computer
                   Vision and Pattern Recognition (CVPR)},
  title =         {LEAD: Minimizing Learner-Expert Asymmetry in
                   End-to-End Driving},
  year =          {2026},
  url =           {https://arxiv.org/abs/2512.20563},
}

@inproceedings{kirby2026driving,
  author =        {Kirby, Ellington and Boulch, Alexandre and Xu, Yihong and
                   Yin, Yuan and Puy, Gilles and Zablocki, {\'E}loi and
                   Bursuc, Andrei and Gidaris, Spyros and Marlet, Renaud and
                   Bartoccioni, Florent and Cao, Anh-Quan and
                   Samet, Nermin and Vu, Tuan-Hung and Cord, Matthieu},
  booktitle =     {Proceedings of the IEEE/CVF Conference on Computer
                   Vision and Pattern Recognition (CVPR)},
  pages =         {32058--32069},
  title =         {Driving on Registers},
  year =          {2026},
}

@article{Li2025GTRS,
  author =        {Li, Zhenxin and Yao, Wenhao and Wang, Zi and
                   Sun, Xinglong and Chen, Joshua and Chang, Nadine and
                   Shen, Maying and Wu, Zuxuan and Lan, Shiyi and
                   Alvarez, Jose M.},
  journal =       {arXiv preprint arXiv:2506.06664},
  title =         {Generalized Trajectory Scoring for End-to-End
                   Multimodal Planning},
  year =          {2025},
  url =           {https://arxiv.org/abs/2506.06664},
}

@inproceedings{feng2026rap,
  author =        {Feng, Lan and Gao, Yang and Zablocki, Eloi and
                   Li, Quanyi and Li, Wuyang and Liu, Sichao and
                   Cord, Matthieu and Alahi, Alexandre},
  booktitle =     {International Conference on Learning Representations},
  title =         {{RAP}: 3D Rasterization Augmented End-to-End
                   Planning},
  year =          {2026},
  url =           {https://arxiv.org/abs/2510.04333},
}

@article{li2026mtdrive,
  author  = {Li, Xidong and Guo, Mingyu and Xu, Chenchao and Li, Bailin and Zhu, Wenjing and Zou, Yangang and Chen, Rui and Wang, Zehuan},
  journal = {arXiv preprint arXiv:2601.22930},
  title   = {{MTDrive}: Multi-Turn Interactive Reinforcement Learning for Autonomous Driving},
  year    = {2026},
  url     = {https://arxiv.org/abs/2601.22930}
}

@inproceedings{navsim,
  author =        {Dauner, Daniel and Hallgarten, Marcel and Li, Tianyu and
                   Weng, Xinshuo and Huang, Zhiyu and Yang, Zetong and
                   Li, Hongyang and Gilitschenski, Igor and
                   Ivanovic, Boris and Pavone, Marco and Geiger, Andreas and
                   Chitta, Kashyap},
  booktitle =     {Advances in Neural Information Processing Systems
                   (NeurIPS) Datasets and Benchmarks Track},
  title =         {{NAVSIM}: Data-Driven Non-Reactive Autonomous Vehicle
                   Simulation and Benchmarking},
  year =          {2024},
  url =           {https://arxiv.org/abs/2406.15349},
}

@inproceedings{chen2021wor,
  author =        {Chen, Dian and Koltun, Vladlen and
                   Kr{\"a}henb{\"u}hl, Philipp},
  booktitle =     {Proceedings of the IEEE/CVF International Conference
                   on Computer Vision (ICCV)},
  pages =         {15590--15599},
  title =         {Learning to Drive from a World on Rails},
  year =          {2021},
}

@article{RoadSafe365,
  author  = {Liu, Xinyu and Jacob, Darryl C. and Liu, Yuxin and Du, Xinsong and Ye, Muchao and Zhou, Bolei and He, Pan},
  journal = {arXiv preprint arXiv:2602.07212},
  title   = {Understanding Real-World Traffic Safety through {RoadSafe365} Benchmark},
  year    = {2026},
  url     = {https://arxiv.org/abs/2602.07212}
}

@misc{sinha2024realtimeanomalydetectionreactive,
      title={Real-Time Anomaly Detection and Reactive Planning with Large Language Models}, 
      author={Rohan Sinha and Amine Elhafsi and Christopher Agia and Matthew Foutter and Edward Schmerling and Marco Pavone},
      year={2024},
      eprint={2407.08735},
      archivePrefix={arXiv},
      primaryClass={cs.RO},
      url={https://arxiv.org/abs/2407.08735}, 
}

@misc{chakraborty2024systemlevelsafetymonitoringrecovery,
      title={System-Level Safety Monitoring and Recovery for Perception Failures in Autonomous Vehicles}, 
      author={Kaustav Chakraborty and Zeyuan Feng and Sushant Veer and Apoorva Sharma and Boris Ivanovic and Marco Pavone and Somil Bansal},
      year={2024},
      eprint={2409.17630},
      archivePrefix={arXiv},
      primaryClass={cs.RO},
      url={https://arxiv.org/abs/2409.17630}, 
}

@misc{anthropic2026opus5,
  author       = {{Anthropic}},
  title        = {{Claude Opus 5} System Card},
  year         = {2026},
  howpublished = {\url{https://www-cdn.anthropic.com/b514064af1408018e64b1ad24e7d5e75850b4ffd/Claude\%20Opus\%205\%20System\%20Card.pdf}},
  note         = {Accessed September 2026}
}

@misc{openai2026gpt6astra,
  author       = {{OpenAI}},
  title        = {Introducing {GPT-6 Astra}},
  year         = {2026},
  howpublished = {\url{https://openai.com/index/gpt-6-astra/}},
  note         = {Accessed September 2026}
}

@misc{qwen2026qwen35,
    title  = {{Qwen3.5}: Towards Native Multimodal Agents},
    author = {{Qwen Team}},
    month  = {February},
    year   = {2026},
    url    = {https://qwen.ai/blog?id=qwen3.5}
}

@misc{zhou2026verifine,
  title={VeriFine: Scaling Verification for Self-Improvement in Embodied Reasoning},
  author={Zhou, Zewei and Luo, Rachel and Cao, Yulong and Xiao, Chaowei and Peng, Chensheng and Li, Boyi and Tian, Thomas and Lian, Zheng and Wang, Yan and Ma, Jiaqi and Ivanovic, Boris and Pavone, Marco and Ding, Wenhao},
  year={2026},
  eprint={2610.08761},
  archivePrefix={arXiv},
  primaryClass={cs.AI},
  url={https://arxiv.org/abs/2610.08761}
}

@article{zou2025diffusiondrivev2,
  title={Diffusiondrivev2: Reinforcement learning-constrained truncated diffusion modeling in end-to-end autonomous driving},
  author={Zou, Jialv and Chen, Shaoyu and Liao, Bencheng and Zheng, Zhiyu and Song, Yuehao and Zhang, Lefei and Zhang, Qian and Liu, Wenyu and Wang, Xinggang},
  journal={arXiv preprint arXiv:2512.07745},
  year={2025}
}
